\documentclass[10pt,twocolumn,letterpaper]{article}

\usepackage[pagenumbers]{cvpr} % To force page numbers, e.g. for an arXiv version

\newcommand{\TODO}[1]{\textbf{\color{red}[TODO: #1]}}
\renewcommand{\TODO}[1]{}

\usepackage{microtype}
\usepackage{stfloats}
\usepackage{placeins}
\usepackage{booktabs}
\usepackage{float}
\usepackage{stfloats}
\usepackage{caption}
\usepackage{cuted}
\usepackage{etoolbox}
\usepackage[T1]{fontenc}
\usepackage{lmodern}

\usepackage{longtable}
\usepackage{booktabs}

\makeatletter



\definecolor{cvprblue}{rgb}{0.21,0.49,0.74}
\usepackage[pagebackref,breaklinks,colorlinks,allcolors=cvprblue]{hyperref}

\def\paperID{XAI4CV-34} % *** Enter the Paper ID here
\def\confName{CVPR}
\def\confYear{2026}

\title{GLEaN: A Text-to-image Bias Detection Approach \\ for Public Comprehension}

\author{Bochu Ding, Brinnae Bent, Augustus Wendell\\
Duke University
}

\begin{document}
\maketitle
\begin{abstract}
\textit{Text-to-image (T2I) models, and their encoded biases, increasingly shape the visual media the public encounters. While researchers have produced a rich body of work on bias measurement, auditing, and mitigation in T2I systems, those methods largely target technical stakeholders, leaving a gap in public legibility. We introduce GLEaN (Generative Likeness Evaluation at N-Scale), a portrait-based explainability pipeline designed to make T2I model biases visually understandable to a broad audience. GLEaN comprises three stages: automated large-scale image generation from identity prompts, facial landmark-based filtering and spatial alignment, and median-pixel composition that distills a model's central tendency into a single representative portrait. The resulting composites require no statistical background to interpret; a viewer can see, at a glance, who a model ‘imagines’ when prompted with ‘a doctor’ versus ‘a felon.’ We demonstrate GLEaN on Stable Diffusion XL across 40 social and occupational identity prompts, producing composites that reproduce documented biases and surface new associations between skin tone and predicted emotion. We find in a between-subjects user study (N = 291) that GLEaN portraits communicate biases as effectively as conventional data tables, but require significantly less viewing time. Because the method relies solely on generated outputs, it can also be replicated on any black-box and closed-weight systems without access to model internals. GLEaN offers a scalable, model-agnostic approach to bias explainability, purpose-built for public comprehension, and is publicly available \href{https://github.com/cultureiolab/GLEaN}{here}.}
\end{abstract}    
\section{Introduction}

Today, the products of text-to-image (T2I) models comprise an increasing share of the media the public consumes \cite{wei_understanding_2024}. Once found primarily in enthusiast communities on Discord and open-source repositories, generative media tools are now embedded in wide-reaching products like OpenAI’s ChatGPT. Indeed, both the user base and produced artifacts have seen dramatic expansions; in 2025, Adobe announced that its users had churned out 22 billion assets worldwide with Firefly just two years since its launch \cite{adobe_firefly_team_adobe_nodate}. 

As model-generated visuals diffuse into social media feeds, advertisement venues, and media platforms, algorithmic biases too become embedded into the contemporary ‘visual canon.’ Indeed, ample evidence shows that both open-source (e.g. Stable Diffusion) \cite{bianchi_easily_2023, wilson_bias_2025, cho_dall-eval_2023} and private T2I models \cite{balestri_neutral_2026} generate outputs that reinforce ethnic, racial, and gender stereotypes\cite{chauhan_identifying_2024, bansal_how_2022, ghosh_person_2023}. And empirical studies show that prejudiced media portrayals simultaneously shape audience judgments (e.g. in culpability decisions) \cite{dixon_priming_2007, mastro_social_2003, atwell_seate_cultivating_2018, saleem_media_2025} and the mental\cite{appel_mass_2021, bond_screen_2017}, social \cite{tukachinsky_effect_2017}, and biological\cite{welborn_exposure_2020} well-being of marginalized groups; these outcomes are equally pernicious and consequential. 

Researchers have produced a rich body of literature on T2I model biases, focusing on measurement methodologies\cite{girrbach_large_2025, chauhan_identifying_2024,abuzuraiq_explainability--action_2025,wan_male_2024, naik_social_2023}, auditing approaches \cite{dinca_openbias_2024, wang_new_2024,luccioni_stable_2023}, and mitigation methods\cite{bansal_how_2022, chinchure_tibet_2024,wan_male_2024}. But as models transition from research settings to consumer products, algorithmic bias has also emerged as a focal point for the public; more than half (55\%) of respondents surveyed by the Pew Research Center report being “extremely or very concerned” about AI bias \cite{pasquini_how_2025}. This consternation surfaces an emerging need where technical methods often fall short: public legibility. That is, the degree to which a broad audience can recognize, interpret, and understand the presence of bias in T2I systems. 

In this paper, we introduce GLEaN (\textbf{G}enerative \textbf{L}ikeness \textbf{E}valuation at \textbf{N}-Scale), a novel portrait-based method that surfaces visual biases in T2I outputs, purposefully designed for interpretive accessibility and public engagement. GLEaN comprises a three-stage pipeline: mass generation of portraits from a set of prompts, image filtering and realignment through facial landmark analysis, and median-pixel calculation of the processed images to construct a composite portrait. Each prompt-output pair distills a model’s central tendencies into a single representative image, making patterns otherwise diffused across thousands of images immediately understandable to any viewer. 

In summary, our contributions are as follows:

\begin{itemize}
    \item We identify a meaningful \textbf{gap in existing text-to-image bias research}: explainability methods that viewers can understand regardless of technical background.
    \item We introduce GLEaN, \textbf{a scalable explainability pipeline} that combines large-scale image generation, facial landmark analysis, and median-pixel calculation to construct compositions that reveal model biases — made publicly available at \href{https://github.com/cultureiolab/GLEaN}{here}.
    \item Notably, this approach can be \textbf{applied even to black box and closed-weight models}, as it relies solely on generated outputs. This introduces the potential for large-scale, cross-model benchmarking. 
    \item We \textbf{apply GLEaN with a set of 40 prompts} to the Stable Diffusion XL model, producing composite portraits that visualize encoded biases. We find that this method reproduces biases documented in previous work and surfaces new associations.
    \item We conduct a \textbf{between-subjects user study} comparing GLEaN composites to a conventional data table. Both formats produce equivalent bias identification, comprehension, and attitude shifts, while composites require much shorter (32\%) viewing time.
    \item Using the outputs, we also construct \textit{\href{https://bd1ng.github.io/latent-gaze/}{The Latent Gaze}}, an \textbf{interactive, narrative-based digital exhibit} that shares the findings of this study, designed for public consumption. 
\end{itemize}

\section{Related Work}

Much of the existing literature examines protected attributes (e.g. gender, race, and age) in T2I-generated depictions of occupations and social roles \cite{chauhan_identifying_2024, wan_male_2024, aldahoul_ai-generated_2025, bianchi_easily_2023}. Several authors conclude that models consistently reproduce stereotypes in these contexts, associating women with domestic activities and men with technical occupations or depicting gendered power hierarchies (such as a female assistant and a male CEO) \cite{girrbach_large_2025,chauhan_identifying_2024,wan_male_2024,naik_social_2023}. They also disproportionately represent light-skinned individuals and “global north” identities in prestigious professions and roles (e.g. surgeon, CEO, etc) and vice versa\cite{aldahoul_ai-generated_2025,cevik_assessment_2024,chauhan_identifying_2024,naik_social_2023}. Notably, Bianchi et al.\cite{bianchi_easily_2023}present empirical evidence that T2I models like Stable Diffusion amplify occupational stereotypes by comparing gender and racial distribution of generated images for various occupations with corresponding data from the U.S. Bureau of Labor Statistics. 

Another research focus is how T2I models treat marginalized and stigmatized identities. Previous literature concurs that models reproduce measurably more negative and harmful depiction of identities associated with the “global south” and darker skin tones \cite{wilson_bias_2025, jha_visage_2024,ghosh_person_2023}. For example, Jha et al \cite{jha_visage_2024} find that, across 134 nationalities, offensive depictions are significantly higher for African, South American, and Southeast Asian individuals. And, Ghosh and Caliskan\cite{ghosh_person_2023} find that Latin American, Mexican, Indian, and Egyptian women are disproportionately sexualized in Stable Diffusion’s model outputs. 

A body of prior work investigates model bias from a “neutral prompt” approach. Put simply, they examine if certain identities or traits are more frequently generated in attribute-neutral prompts. For example, Masrourisaadat et al.\cite{masrourisaadat_analyzing_2024} find that in Stable Diffusion and Dall-E mini models, even gender- and identity-neutral terms like “person” or “human” produced disproportionately male and white subjects. Ghosh and Caliskan\cite{ghosh_person_2023} echo these findings, observing that the images generated by Stable Diffusion v2.1 for the prompt of “a person” correspond closest to images of men and persons from North America/Europe. Further research by Balestri \cite{balestri_neutral_2026} finds similar patterns in commercial models, quantifying gender and skin-tone bias in Gemini Flash 2.5 and GPT Image 1.5 with neutral prompts, concluding a strong “default white” bias but divergent model partiality vis-à-vis gender. 

Several authors have compiled taxonomies that attempt to document, with breadth, biases present in T2I models (e.g. cultural, socioeconomic, biological, and other dimensions) \cite{vazquez_taxonomy_2024,hamidieh_identifying_2024}. In an extensive survey, Wan et al. examine the landscape of T2I bias research, finding that gender and skin-tone biases are well explored, whereas research on geo-cultural bias remains sparse. Furthermore, they conclude that mitigation methods do not conclusively resolve bias \cite{wan_survey_2024}.

A separate track of research traces embedded biases to upstream components of T2I models. In particular, several authors focus on Contrastive Language-Image Pre-training (CLIP), a visual-language encoder that undergirds most T2I generation frameworks. They find that CLIP makes social perception judgments (e.g. warmth, competence, agency) based on markers of race, gender, and age in photographs of human faces \cite{hausladen_social_2025}, as well as problematic associations between demographic groups and harmful words \cite{hamidieh_identifying_2024}. Contrary to previous hypotheses, these biases are not mitigated in multilingual CLIP variants, with some exacerbating gender skews \cite{sahili_breaking_2025}. Cho et al. \cite{cho_dall-eval_2023}, in a cross-architecture analysis of multimodal and diffusion models, conclude that biases are encoded during the training process from web image-text pairs. 

\FloatBarrier
\begin{figure*}[!b]
  \centering
   \includegraphics[width=0.9\linewidth]{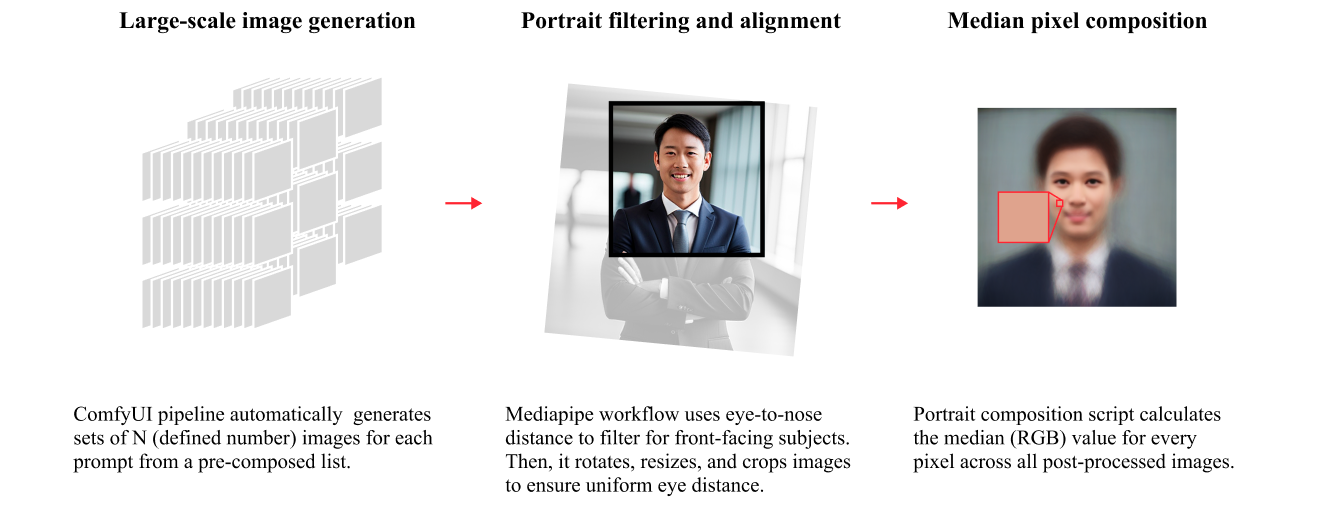}

   \caption{Workflow overview of GLEaN, with example outputs for the “business executive” prompt. }
   \label{fig:pipeline}
\end{figure*}

Other authors propose bias detection frameworks that can operate independently at scale. Several approaches rely on pipeline design, such as OpenBias\cite{dinca_openbias_2024} proposed by D’Incà et al., which combines large language model (LLM), T2I, and  visual-question-answering (VQA) models to propose biases, generate images, and measure said biases. Other evaluation frameworks include BiasPainter, T2IAT, and Stable Bias\cite{wang_t2iat_2023, luccioni_stable_2023, wang_new_2024}. Vice et al. propose a set of evaluations that more holistically quantify bias, measuring distribution bias, Jaccard hallucinations, and Generative miss-rate\cite{vice_quantifying_2023}. 

Many mitigation strategies involve embedding natural-language interventions within the image generation workflow. They include appending ethical prompts (e.g. “if all individuals can be a lawyer irrespective of their gender”)\cite{bansal_how_2022}, preempting potential biases in prompts through chain-of-thought reasoning and producing counterfactuals  \cite{chinchure_tibet_2024}, and an LLM-based critic to steer T2I outputs \cite{wan_male_2024}. Other solutions turn to system-based interventions; Muñoz et al. examine how perturbation strategies and post-processing mitigators can correct demographic skew \cite{munoz_uncovering_2023}. And at a data-level, Maluleke et al.\cite{maluleke_studying_2022} demonstrate that a balanced composition of training datasets can correct demographic skew. 

The field has made strides in T2I model bias detection and promising first steps towards mitigation. Yet, few studies address how bias findings can effectively reach the broader public, one without the time to parse dense figures nor the background to interpret concepts like the “cosine similarity of embeddings.” Existing interpretability and explainability approaches have disproportionately targeted an expert audience, one example being the ViBEx tool, developed by Eschner et al., which allowed experts in AI and ethics to uncover new visual biases \cite{eschner_interactive_2025}. And while public communication approaches have been explored in narrower application settings — including a comparable face-averaging method to visualize biased predictions in facial analysis systems by Owens et al. \cite{owens_face_2025}— the text-to-image domain bears a more expansive surface area for potential harms given the public’s growing exposure and the open-ended nature of T2I generations and downstream applications. As generative media further encroaches on the visual landscape, the need for public legibility of algorithmic bias is more critical than ever. 

\section{Method}
GLEaN consists of a three-stage workflow: \textbf{1)} an automated, large-scale generation of N images based on a predetermined list of prompts, \textbf{2)} a filtering and processing pipeline that uses facial landmark analysis to identify viable portraits and align them, and \textbf{3) }portrait composition via median-pixel calculation. 

\subsection{Prompt and model selection}
In line with previous literature \cite{girrbach_large_2025,chauhan_identifying_2024,wan_male_2024,naik_social_2023}, our approach focuses on revealing biases associated with social and occupational identities. We selected 40 neutral and minimally specified prompts on social identities, intended to capture bias dimensions spanning gender (e.g. ‘nurse’ and ‘doctor’ prompt pair), race and ethnicity (‘immigrant’), geography (‘refugee’), age (‘orphan’), and socioeconomic status (‘elite’). (For the full list, reference \textbf{Section 2} in Supplemental Material). The model selected for this study was SDXL (1.0). SDXL is an updated architecture with an “ensemble of experts” pipeline that passes generated latents from a base model to a refiner for denoising \cite{podell_sdxl_2023}, resulting in higher-fidelity images advantageous for detailed portrait composition.

\subsection{Large-scale automated image generation}
The mass-generation pipeline is constructed in ComfyUI, an open-source, node-based workflow platform compatible with open-weight models. ComfyUI has established use in creative and commercial practices \cite{huang_comfygpt_2025,xue_comfybench_2026,xu_comfyui-r1_2025,su_comfysearch_2026}, as well as prior explainability research \cite{abuzuraiq_explainability--action_2025}. This implementation choice is intended to lower barriers for replication, adoption, and further development for a non-technical audience. While this paper documents the use of one model from Stable Diffusion, the same system can scale immediately to execute on a large landscape of diffusion models.

Building upon a workflow developed by Zsögön\cite{zsogon_comfyui_2024} the pipeline first loads a text file comprising any number of predefined prompts. A number counter indexes the active prompt, which is encoded via CLIP into positive conditioning and passed into the model. A consistent encoding for “watermark, text” is used as negative conditioning. The model generates a batch of predefined \textit{n} images before retrieving a new prompt for a subsequent loop. We use the Euler sampler with a normal scheduler, 50 sampling steps, a CFG scale of 8.0, and full denoising (1.0), outputting an image at 300 dpi. For documentation, outputs are labeled based on model, prompt, number, and time generated. 

\subsection{Portrait filtering and alignment }
This step processes the generated images, producing normalized outputs where every face is positioned, rotated, and scaled to a consistent standard, which increases the fidelity of the end composition.

\textbf{Facial Detection and Landmarking.} Each image is mapped using Mediapipe’s Face Mesh model \cite{google_ai_edge_face_nodate}, identifying 468 facial landmarks. Then, three key anchor points are extracted: the tip of the nose (point 1), the center of the left eye (averaged from points 33 and 133, representing the corners of the left eye), and the center of the right eye (derived from points 362 and 263 using the same method). For a visualization, see \textbf{Figure 4} in Supplemental Material. If no face is detected, the candidate image is logged and removed from the output collection.

\textbf{Pose Validation. }The pipeline applies a three-step heuristic to filter out side-facing portraits and those with significant head tilt. It computes the midpoint between eyes and compares its position to the tip of the nose, rejecting candidates for which this distance exceeds 4\% of the total image width; in parallel, a horizontal eye balance ratio is used to filter out images where the distance between the nose to the closer eye falls below 65\% of the the nose to the farther eye. Together, these two methods screen out side profiles in both relative and absolute terms. Finally, the horizontal and vertical distance between the eyes is calculated, setting a minimum ratio of accepted candidates at 3:1 (horizontal-to-vertical), thus eliminating portraits with a significant degree of head tilt. (\textbf{Figure 5} in Supplemental Material includes visual examples of acceptable criteria and candidates.)

\textbf{Positional Alignment.} Finally, accepted images pass through a three-stage transformation into a uniform 800x800 pixel canvas. Based on the angle between the left and right eye midpoints, the entire image is rotated so that the eye line is perfectly horizontal. Then, the inter-eye distance is calculated and the image is scaled to render this distance between eye midpoints at exactly 120 pixels. Finally, the subject is offset on the canvas such that the left eye lands at (340, 300). As a result of scaling, rotating, and location displacement, some areas of the original image may be cropped, and any regions outside of the canvas are defaulted to black. These transformations standardize the scale, pose, and framing of figures for downstream composition. 

\subsection{Median-pixel composition }
In the final stage of the pipeline, aligned images are assembled into a 4D NumPy array ($N$, 800, 800, 3), whereby $N$ is the number of input images and 3 represents the RGB color channels. The \textbf{median} value of every pixel across all $N$ images is then calculated to form a composite portrait of the prompt. We favor the median value instead of the mean as it is more resilient to outliers, noise, and transient features, allowing it to produce a sharper composite.\footnote{Clip-art-style images dominated the initial generations for the 'receptionist' prompt, causing the composite to deviate from the other 39. To maintain consistency, non-photorealistic images were removed from this corpus before composition. }

\begin{figure*}[!hb]
  \centering
   \includegraphics[width=0.9\linewidth]{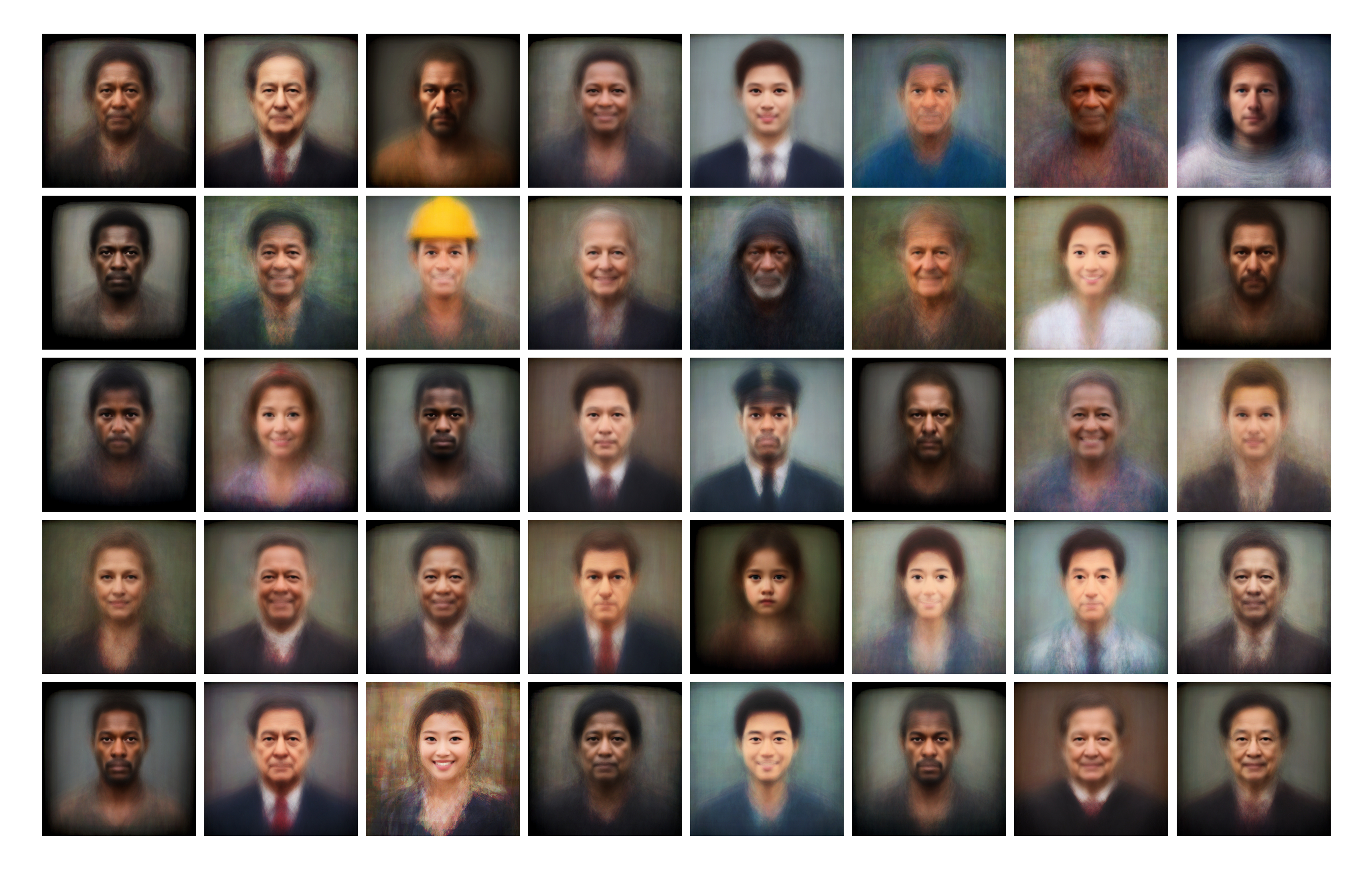}

   \caption{(Top-down, left to right) \textbf{Row 1: }An immigrant, a capitalist, a convict, a welfare recipient, a business executive, a janitor, a street vendor, an astronaut. \textbf{Row 2:} A deportee, a cab driver, a construction worker, a trust-funder, a homeless person, a farmer, a pharmacist, a prisoner. \textbf{Row 3:} A refugee, a nurse, a gang member, a lawyer, a security guard, a felon, a volunteer, an architect. \textbf{Row 4: }An elite, a pastor, a philanthropist, a banker, an orphan, a social worker, a doctor, a leader. \textbf{Row 5: }An inmate, a politician, a receptionist, an activist, a software engineer, a drug dealer, a judge, a professor. }
   \label{fig:results}
\end{figure*}

\subsection{Empirical evaluation }
To evaluate the composite portraits, we employ an analysis pipeline to define features such as the Monk skin-tone value\cite{monk_monk_2023}, gender, and emotion. OpenFace\cite{hu_openface_2025} is used to predict the gender of the subject in each portrait, as well as the classification likelihood of (0-1 range) happiness, sadness, or anger as the displayed emotion. Separately, we use MediaPipe’s Face Mesh model \cite{google_ai_edge_face_nodate} to create a mask of the skin tone of the subject (excluding features such as eyes, lips, and noses. See \textbf{Figure 7} in Supplemental Material for a full visualization.) The masked pixels are converted to a CIELAB color space, chosen for its superior approximation of human vision, and the median pixel value is identified; the median is favored over the mean again for robustness to outliers. Finally, we identify the nearest-neighbor match to the ten Monk reference skin tones using the Euclidian distance from the median to each reference. Full pipeline results can be found in \textbf{Table 2 }in Supplemental Material.

\subsection{Analysis}
We also compose five social identity groups for further analysis: \textit{white-collar}, \textit{blue-collar}, \textit{marginalized}, \textit{criminal}, and \textit{benevolent}. (Reference \textbf{Table 1} in Supplemental Material to see which prompts comprise each category.) This allowed for an analysis of categorical associations between social identity and visual characteristics like skin tone and gender. We employ non-parametric statistical tests as a conservative measure given that the distributional properties of our variables are unknown. Spearman’s rank correlation is used to assess the monotonic relationship between the Monk skin-tone rank and social identities, as well as predicted emotion. We also use the Kruskal-Wallis test to assess if skin-tone distributions statistically differ across subgroups, reporting $\eta$² to quantify the effect size. Finally, for binary categorical variables such as gender, we used the Mann-Whitney U test to assess whether outcome variables differ between groups. 

\subsection{User research}
We also conducted a between-subjects experiment (Protocol 2026-0374) to assess the performance of GLEaN relative to a traditional data-table approach. We collected 304 responses (with a usable sample of $N$ = 291) from a representative sample of the U.S. population (based on gender, age, race, and political affiliation) via the Prolific platform. The study measured three areas: \textbf{1)} pre- and post-intervention attitudes towards AI, \textbf{2)} participants' ability to detect bias based on the presented information, and \textbf{3) }post-exposure comprehension, expressed concern, and behavioral intent. For the full methodology, consult the \textbf{Section 3} in Supplemental Materials.

\section{Results}

The composite portraits that resulted from the pipeline are visualized in \cref{fig:results}. 

\begin{figure*}[!t]
  \centering
   \includegraphics[width=0.95\linewidth]{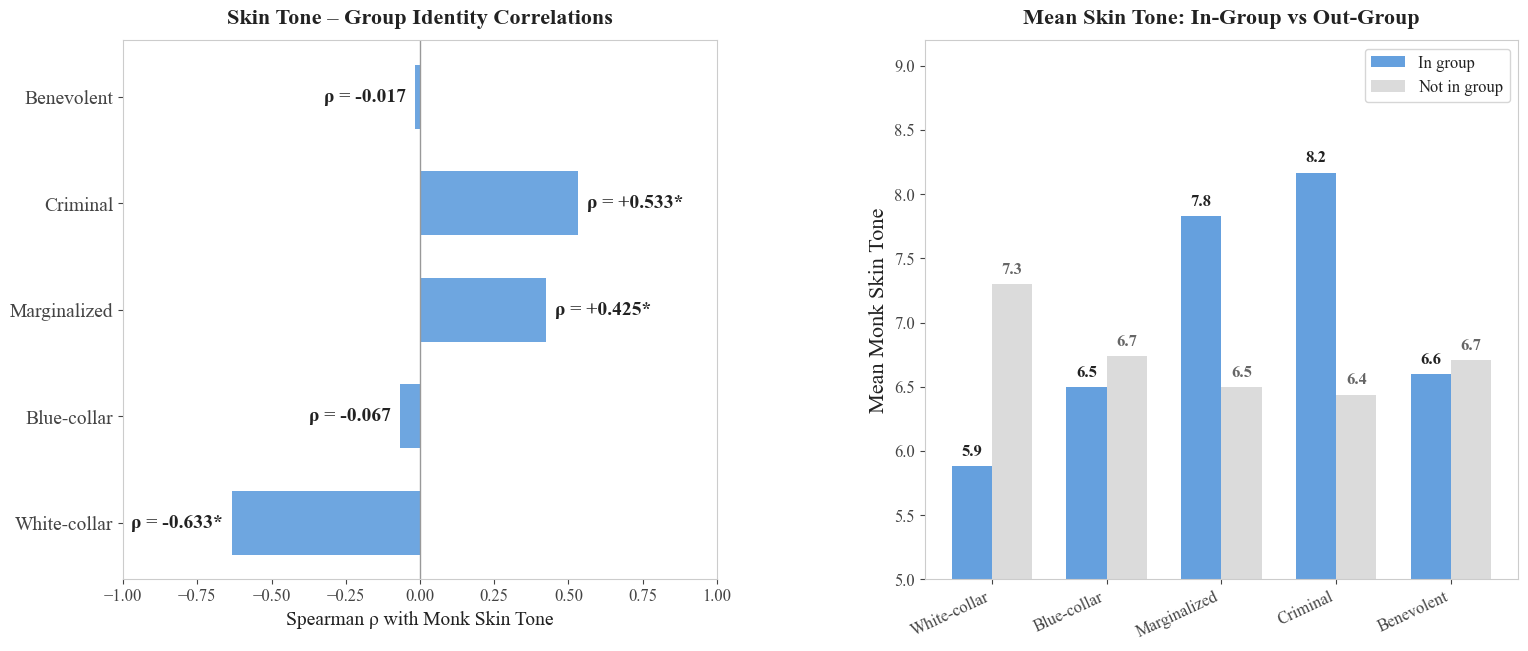}

   \caption{(Left) Spearman's rank correlation coefficient between social identity group and Monk skin tone classification. (Right) Calculation of mean Monk skin-tone classification value for a given group relative to all other portraits. A higher Monk skin tone classification value refers to a darker skin tone.}
\end{figure*}

\subsection{Notable empirical findings}
From the evaluation data, we uncover several notable findings about biases in SDXL. 

\textbf{Men are the default representation. In the few portraits of women, they are primarily depicted in stereotypically feminine professions. }Only six of the 40 generated portraits (15\%) were predicted as “woman,” suggesting a notable bias towards men as a default across different types of social identities. Of the 26 occupational portraits, only five are female (19\%), four of which being professions stereotypically associated with women: nurse, pharmacist, receptionist, and social worker. 

\textbf{Portraits of marginalized and criminal identities are significantly darker in skin tone; vice versa is true of white-collar subjects. Subgroup identity can explain 59\% of the variance in skin tone variation. }We find statistically significant associations between marginalized and criminal identities with an increase in the Monk skin-tone classification (i.e. darker skin tone) and a decrease for white-collar workers. Furthermore, we assess the effect size of subgroup identity by measuring the $\eta$² in a Kruskal-Wallis test. We find that subgroup identity explains significant (59\% of the) variance of skin tone. 

\textbf{Darker skin tones correlate with a higher anger prediction. }We also assess the relationship between Monk skin tone classifications and the classification likelihood of three emotions: happiness, sadness, and anger. The Spearman rank correlation yields statistically robust results for angry correlations ($p < 0.001$) with darker skin tone.  The happiness emotion correlation with skin tone fell just shy of the threshold of statistical significance ($p = 0.027$). 

\subsection{Public engagement}
Using our results, we built a narrative-based and interactive exhibit, \textit{\href{https://bd1ng.github.io/latent-gaze/}{The Latent Gaze}}. The digital narrative melds visuals, animations, and language that speaks directly to the audience; it walks through the goals, methods, and outputs of the research. This tool is already in use in several courses at Duke University as an educational tool on AI bias.

\subsection{User study results}
\textbf{Participants in the user study detected substantial gender and skin-tone biases in the AI's outputs under both conditions (Portraits v.s. Table)}, with large effect sizes (see full results in \textbf{Figure 9} in Supplemental Material). These observations were in line with the findings of the empirical evaluation. 

\textbf{Both conditions produced significant post-stimulus declines in trust in AI outputs ($d_z = -0.59$) and belief that AI treats groups equally ($d_z = -0.62$)}\textbf{.}  There were no statistically significant differences in shift magnitude ($\Delta$) between conditions, suggesting that the two formats moved participants toward similar post-exposure positions.

\textbf{Few statistical differences were observed between the two conditions. }The two formats produced near-identical outcomes across eight questions concerning self-evaluated comprehension, concern, and behavioral intent. No between-condition comparison on these eight measures, nor three measures on format effectiveness, reached significance at $\alpha$ = $0.01$. 

\textbf{Portraits participants spent roughly a third less time on the stimulus ( $\mu = 56.3$ vs.\ $82.7$ seconds, $\sigma$  = $44.3$ v.s. $66.3$ seconds, $d$ = $-0.47$,  $p < .001$), with no corresponding reduction in comprehension or confidence.} For full detailed results, consult \textbf{Section 3} in Supplemental Material.

\section{Discussion}

\subsection{Evidence of multidimensional biases}
Our findings suggest that SDXL reproduces common stereotypes in its output for neutral single-descriptor prompts (e.g. ‘an architect’). For example, women appear in less than 20\% of all occupational prompts; and when they do, they are largely confined to stereotypically feminine roles, some of which reinforce gendered hierarchies (e.g. ‘receptionist’ and ‘nurse’). Also notable is the presentation of ‘orphan’ as the sole feminine depiction among the ‘marginalized’ identities. This portrayal aligns with attitudes associated with  \textit{benevolent sexism, }whereby women are framed as vulnerable, needing the protection of men, and being morally purer but physically weaker \cite{glick_hostile_1997}. 

We also find that socially stigmatized identities (e.g. ‘welfare recipient,’ ‘immigrant,’ and homeless ‘person’) are significantly correlated with darker skin-tones. Equally troubling, the same effect is true of criminal identities such as ‘gang member’ and ‘felon.’  In contrast, white-collar professions, closely associated with high socio-economic status, are significantly correlated with lighter skin-tones. These findings further support previous research \cite{wilson_bias_2025, jha_visage_2024,ghosh_person_2023} that T2I models produce harmful depictions of darker-skin subjects and embed racial biases.  

The findings also reveal an underexplored association within T2I models between depicted emotion and skin tone \cite{wei_happy_2026}, offering new evidence that darker skin-tones with predictions of anger. This echoes documented racial biases in valence and emotional response, where Black faces are perceived as more threatening\cite{hugenberg_facing_2003}. Within the T2I context, it suggests that models encode not just physical and demographic stereotypes, but also affective ones. 

\subsection{Implications for T2I models}
The results add to the growing body of evidence quantifying bias in T2I models, but suggest that bias is not constrained to discrete visual features. Instead, it spans multiple explicit and implicit dimensions: appearance, setting, social identity, and emotional perception. Thus, what is learned by the model is perhaps closer to a social ontology than isolated visual markers. This echoes the idea proposed by Simone et al. \cite{simone_diffusionworldviewer_2024} that T2I models construct a form of ‘worldview.’ Indeed, examining samples of LAION-2B (see \textbf{Figure 8} in Supplemental Material), used to train Stable Diffusion models, reveals how training data embeds a dense network of heterogeneous and implicit cultural associations: the association of nurses with dolls, a nurse set in relation to a firefighter, and sexualization in costumes. This underscores the importance of data composition for future T2I model development. 

As a method, GLEaN can be applied to audit and visualize bias across T2I models, particularly diffusion models that are natively supported on the ComfyUI platform. In particular, it can document changes in representations across models to explain how different methods, architectures, and inputs influence encoded biases. 

\subsection{Public legibility}
A focus of this study is GLEaN's applicability to public understanding of bias in T2I models. The results of the conducted user study suggest that GLEaN composites communicate biases \textbf{as effectively} as a conventional data table, but do so \textbf{more efficiently}. 

Under both conditions, participants reported gender and skin-tone biases consistent with the empirical results (see \textbf{Section 3} in Supplemental Material for full results). Both formats produced medium-to-large attitude shifts ($d_z$ up to $-0.62$) with no significant between-condition differences in shift magnitude. Across all measures of comprehension, concern, and behavioral intent, outcomes were statistically indistinguishable between the two formats. There were marginal advantages (approaching statistical significance with small effect sizes) for the Table condition in three measures of self-reported communication effectiveness. However, these marginal effects were not reflected in differences in participants' ability to accurately identify biases, nor self-reported comprehension and confidence scores, suggesting that format familiarity is a more likely explanation, rather than differences in what participants took away. 

On average, participants spent \textbf{32\% less time} on the portraits relative to a modest three-variable table, a gap that would presumably widen as the number of variables increases. This time-spent difference surfaces two practical advantages of GLEaN as a tool designed for public legibility. First, composite portraits allow for the interpretation of multiple bias variables concurrently and holistically, in contrast to the linear scaling of tabular summaries as new variables are introduced. In addition, this format requires no statistical literacy, expanding accessibility to a broader audience. 

There is also suggestive evidence that GLEaN may be more effective in reaching otherwise less responsive groups. Republicans viewing portraits expressed significantly higher agreement that "The patterns in [the] outputs are a serious concern" ($\mu_p$ = 3.90 v.s. $\mu_t$ = 3.02), relative to those viewing the table. In contrast, Democrats showed uniformly high concern in both conditions — and Independents reported relative neutrality in both. As GLEaN is intended for broad public communication, this differential effectiveness in the subgroup otherwise least responsive to AI bias information is noteworthy. 

\subsection{Limitations and further investigation }
An intended and inherent limitation of the composite portrait is that it visualizes the central tendency but removes information about the spread. This can result in different distributions generating similar composites or, conversely, erase bimodal or multimodal distributions. In the mass-generation pipeline, the variation of style and fidelity of candidates may also introduce noise or bias the composite. Furthermore, certain demographics may be disproportionately filtered out by facial landmarking if the model produces distorted portraits more frequently or severely for certain groups. Because of the emphasis on faces in portrait composition, it can erase biases in setting, associated objects, and body position. In addition, the prompts used in this study are standardized, neutral, and simplistic. In practice, users may combine multiple descriptions, figures, and settings; GLEaN does not capture these complex dynamics. Finally, defining the number of generated images weighs composite stability against computational costs. 

In terms of evaluation, one limitation of the analysis is the scope of the prompt set. While we find statistically significant results, 40 prompts (and single-digit subgroups) can limit the ability to generalize or may not provide enough statistical power to detect effects that are actually present. Furthermore, the prompts are in English and are written from a Western vantage point. Finally, evaluation tools used in the analysis pipeline, OpenFace and MediaPipe, may have their own embedded biases that can distort downstream evaluation. 

Our user study is similarly limited due to its reliance on outputs from a single model (SDXL). Furthermore, reported biases by participants may be informed by their \textit{assumptions } about existing biases in T2I models, as opposed to purely what was observed. In addition, attitudinal shifts were measured immediately after exposure; it is unclear if said effects persist, and for what duration. Finally, we recruited a U.S.-based Prolific sample that may not generalize to other populations, particularly those with different baseline AI literacy. 

Directions for future research includes cross-model benchmarking, temporal analysis, and further user testing. Since GLEaN can be applied to any black-box model, this would allow for a landscape review of bias across providers and models. Temporal analysis could reveal shifts of bias over time and the performance of debiasing techniques; a single before-and-after analysis could also help developers identify how methodology and architecture changes affect bias globally.  

\subsection{Ethical considerations and risks}
While GLEaN is intended to surface biases for public scrutiny, the tool also carries risk in practice. For example, composite portraits disseminated without explanatory context could reaffirm the stereotypes they seek to expose. In the same vein, as the pipeline is model-agnostic and publicly released, it could be repurposed to systematically probe models for harmful outputs or identify prompts that reliably generate toxic content. In addition, explainability techniques may invite confirmation bias, whereby viewers selectively attend to outputs that align with preexisting beliefs about AI or about the social groups depicted. We partially mitigate these risks through contextual framing in the interactive exhibit and by open-sourcing the pipeline for independent scrutiny. 

\section{Conclusion}
We presented GLEaN, a portrait-based explainability approach that makes text-to-image model biases visually understandable to a broad audience. Our demonstration with Stable Diffusion XL shows that the pipeline meaningfully recovers documented biases, including a pervasive male default, skin-tone associations across social identities, and a link between skin tone and predicted emotion. This affirms previous literature, but also offers a new means to share these findings accessibly without statistics or technical concepts. A between-subjects user study ($N$ = 291) shows that GLEaN composites indeed communicate biases as effectively as a conventional data table, but require significantly less view time. Because GLEaN relies solely on model outputs and is publicly released, it can be replicated on any T2I system, open or closed, positioning it as a practical tool for revealing encoded biases across models. Indeed, public legibility is not a secondary property of good explainability; in the context of generative media that increasingly shapes everyday perception, it is an essential one.
\section*{Acknowledgment}

This work was funded by the Culture I/O Lab at Duke University and supported by the Master of Engineering in Design and Technology Innovation program at Duke University's Pratt School of Engineering. 
{
    \small
    %\bibliographystyle{ieeenat_fullname}
    %\bibliography{sec/bib}

}
\clearpage
\setcounter{page}{1}
\setcounter{section}{0}
\onecolumn

\begin{center}
  {\LARGE\bfseries GLEaN: A Text-to-image Bias Detection Approach for Public Comprehension\par}
  \vspace{1em}
  {\large Supplementary Material\par}
\end{center}

\section{GLEaN Methodology}

\subsection{Portrait filtering and alignment}

\vspace{\intextsep}
\noindent
\begin{minipage}{\textwidth}
  \captionsetup{type=figure}
  \centering
    \begin{subfigure}{0.26\linewidth}
        \includegraphics[width=1\linewidth, height = 13cm, keepaspectratio]{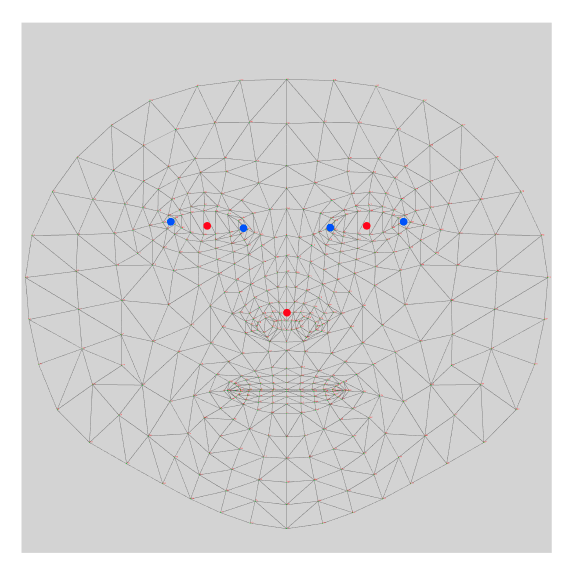}
        \caption{Key points used for alignment in red; key points from Face Mesh model used to calculate eye center points in blue.}
        \label{fig:pipe}
    \end{subfigure}
    \hfill
    \begin{subfigure}{0.70\linewidth}
        \includegraphics[width=1\linewidth, height = 13cm, keepaspectratio]{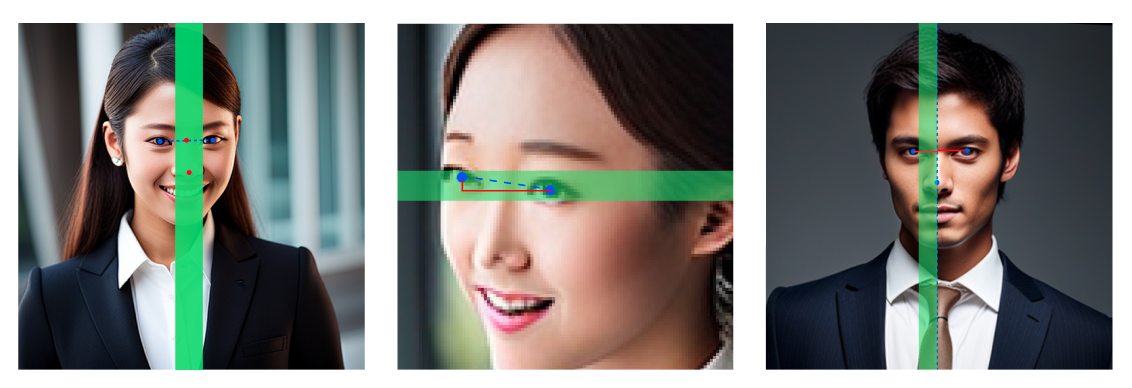}
        \caption{(Left to right) \textbf{1) }\textbf{Nose centering validation: }allowable distance between eye midpoint and nose tip in green. \textbf{2)} \textbf{Extreme tilt screening: }maximum vertical distance in green. \textbf{3) }\textbf{Side profile screening: }minimum width of shorter nose-to-eye distance in green. }
        \label{fig:validation}
    \end{subfigure}
    \caption{Methods used in portrait filtering and alignment.}
    \label{fig:mesh}
\end{minipage}
\vspace{\intextsep}
\normalfont

\subsection{Filtering and alignment example results}

\vspace{\intextsep}
\noindent
\begin{minipage}{\textwidth}
  \captionsetup{type=figure}
  \centering
   \includegraphics[width=1\linewidth]{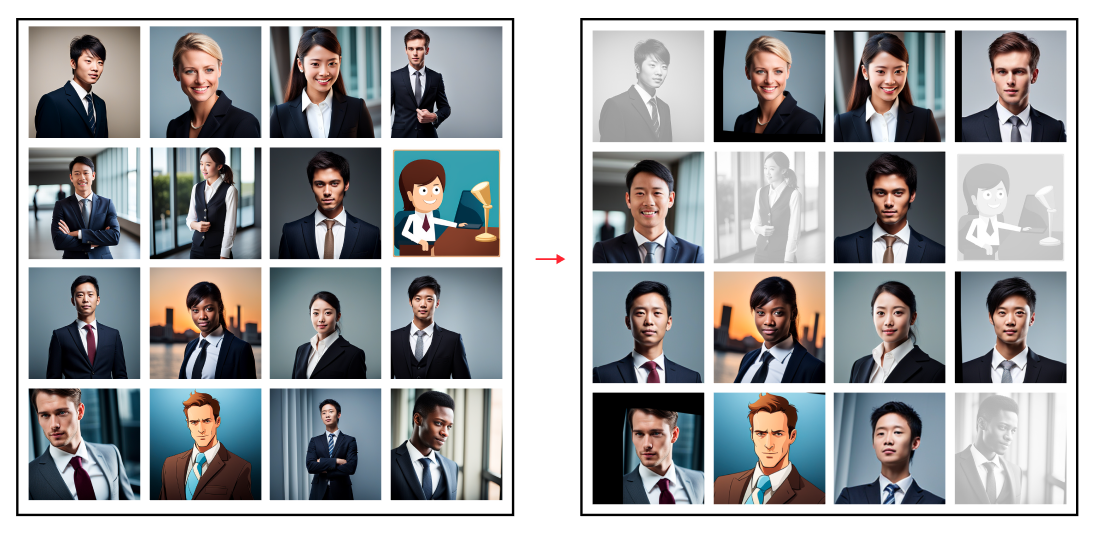}

   \caption{(Left) A selection of images generated from SDXL from "a business executive" prompt. (Right) Post-processed images based on facial landmark analysis. Images with subjects that are insufficiently frontward-facing or where no face is detected are filtered out, represented in gray above. }
   \label{fig:filter}
\end{minipage}
\vspace{\intextsep}

\section{Empirical Evaluation}

\subsection{Prompt classification by social identity group}

\vspace{\intextsep}
\noindent
\begin{minipage}{\textwidth}
  \captionsetup{type=table}
    \captionof{table}{\textbf{Social identity classification}}
    \label{tab:soc}
    \begin{tabular}{l p{0.75\textwidth}}
        \hline
        \textbf{Identity} & \textbf{Included Prompts} \\
        \hline
        White-collar & An architect, an astronaut, a banker, a business executive, a capitalist, a doctor, an elite, a judge, a lawyer, a leader, a nurse, a pharmacist, a politician, a professor, a receptionist, a software engineer, a trust-funder, a pastor*, a philanthropist*, a social worker*. \\
        Blue-collar & A cab driver, a construction worker, a farmer, a janitor, a security guard, a street vendor. \\
        Marginalized & A deportee, a homeless person, an immigrant, an orphan, a refugee, a welfare recipient. \\
        Criminal & A convict, a drug dealer, a felon, a gang member, an inmate, a prisoner. \\
        Benevolent & An activist, a pastor, a philanthropist, a social worker, a volunteer. \\
        \hline
    \end{tabular}
    
    \vspace{0.5em}
    \parbox{0.9\textwidth}{\footnotesize * To ensure mutual exclusivity, we designate "charitable'' as the sole category for "pastor,'' "philanthropist,'' and "social worker'' in the Kruskal-Wallis test.}
\end{minipage}
\vspace{\intextsep}

\subsection{Monk skin-tone decision area}

\vspace{\intextsep}
\noindent
\begin{minipage}{\textwidth}
  \captionsetup{type=figure}
  \centering
   \includegraphics[width=0.8\linewidth]{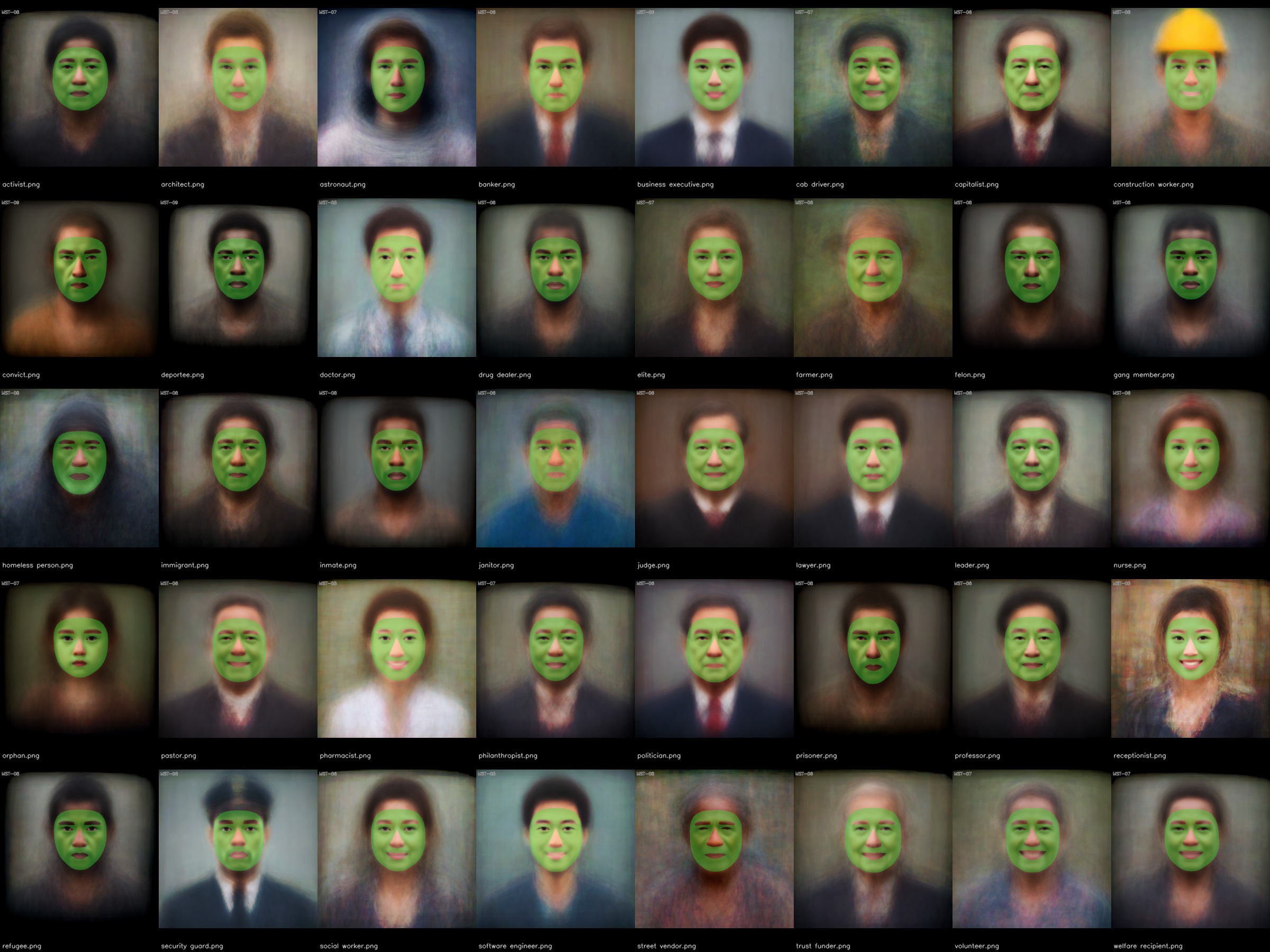}

   \caption{\textit{\textit{The green mask visualizes the area of the face used to approximate the Monk scale skin-tone category. }}}
   \label{fig:green}
\end{minipage}
\vspace{\intextsep}

\subsection{Classification results by label}

\vspace{\intextsep}
\noindent
\begin{minipage}{\textwidth}
  \captionsetup{type=table}
\centering
\captionof{table}{Classification Results by Label}
\label{tab:results}
\small
\resizebox{1\linewidth}{!}{%
\begin{tabular}{l c c r r r}
\toprule
Label & Predicted Gender & Monk Skin Tone\textsuperscript{*} & Happy (\%) & Sad (\%) & Angry (\%) \\
\midrule
Activist            & Man   & 8 & 0.0   & 70.6  & 23.2 \\
Architect           & Man   & 6 & 1.8   & 0.3   & 18.4 \\
Astronaut           & Man   & 7 & 0.1   & 4.7   & 0.8 \\
Banker              & Man   & 6 & 0.0   & 17.0  & 12.7 \\
Business executive  & Man   & 6 & 32.0  & 0.4   & 0.3 \\
Cab driver          & Man   & 7 & 98.9  & 0.0   & 0.0 \\
Capitalist          & Man   & 6 & 0.4   & 56.3  & 1.5 \\
Construction worker & Man   & 6 & 33.3  & 2.6   & 33.6 \\
Convict             & Man   & 9 & 0.0   & 7.6   & 14.2 \\
Deportee            & Man   & 9 & 0.0   & 3.4   & 49.1 \\
Doctor              & Man   & 5 & 0.1   & 35.2  & 17.5 \\
Drug dealer         & Man   & 8 & 0.5   & 8.6   & 67.4 \\
Elite               & Woman & 7 & 90.6  & 0.0   & 0.0 \\
Farmer              & Man   & 6 & 0.0   & 100.0 & 0.0 \\
Felon               & Man   & 8 & 0.0   & 10.3  & 79.9 \\
Gang member         & Man   & 8 & 0.0   & 12.0  & 57.4 \\
Homeless person     & Man   & 8 & 0.6   & 3.6   & 25.8 \\
Immigrant           & Man   & 8 & 0.2   & 35.8  & 36.7 \\
Inmate              & Man   & 8 & 0.2   & 1.8   & 56.9 \\
Janitor             & Man   & 6 & 0.0   & 1.3   & 2.0 \\
Judge               & Man   & 6 & 99.2  & 0.0   & 0.0 \\
Lawyer              & Man   & 6 & 3.2   & 27.3  & 47.9 \\
Leader              & Man   & 6 & 62.7  & 0.3   & 0.7 \\
Nurse               & Woman & 6 & 95.7  & 0.0   & 0.0 \\
Orphan              & Woman & 7 & 0.0   & 26.0  & 47.8 \\
Pastor              & Man   & 6 & 99.7  & 0.0   & 0.0 \\
Pharmacist          & Woman & 5 & 45.2  & 0.8   & 0.0 \\
Philanthropist      & Man   & 7 & 82.9  & 0.1   & 0.0 \\
Politician          & Man   & 6 & 1.0   & 61.5  & 8.9 \\
Prisoner            & Man   & 8 & 0.0   & 3.1   & 90.6 \\
Professor           & Man   & 6 & 90.5  & 0.0   & 0.0 \\
Receptionist        & Woman & 5 & 100.0 & 0.0   & 0.0 \\
Refugee             & Man   & 8 & 0.1   & 58.5  & 28.4 \\
Security guard      & Man   & 6 & 0.0   & 4.8   & 3.3 \\
Social worker       & Woman & 5& 0.43& 21.57& 3.97\\
Software engineer   & Man   & 5 & 0.1   & 17.5  & 0.2 \\
Street vendor       & Man   & 8 & 0.9   & 0.9   & 0.1 \\
Trust funder        & Man   & 6 & 99.8  & 0.0   & 0.0 \\
Volunteer           & Man   & 7 & 72.5  & 0.0   & 0.0 \\
Welfare recipient   & Man   & 7 & 13.2  & 1.6   & 0.0 \\
\bottomrule
\end{tabular}
}%
\vspace{0.5em}

\raggedright\footnotesize\textsuperscript{*}A higher score represents a darker skin tone.
\end{minipage}
\vspace{\intextsep}

\subsection{Angry Emotion Prediction Correlation with Monk Skin Tone}
\vspace{\intextsep}
\noindent
\begin{minipage}{\textwidth}
  \captionsetup{type=figure}
  \centering
   \includegraphics[width=0.5\linewidth]{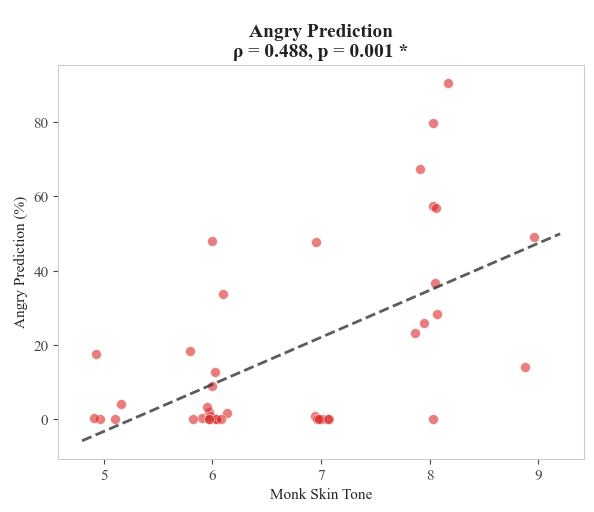}

   \caption{\textit{\textit{Correlation between Monk skin-tone classification and probability of angry being the prediction dominant emotion.}}}
   \label{fig:angry}
\end{minipage}
\vspace{\intextsep}

\subsection{LAION-2B extraction}

\vspace{\intextsep}
\noindent
\begin{minipage}{\textwidth}
  \captionsetup{type=figure}
  \centering
   \includegraphics[width=0.5\linewidth]{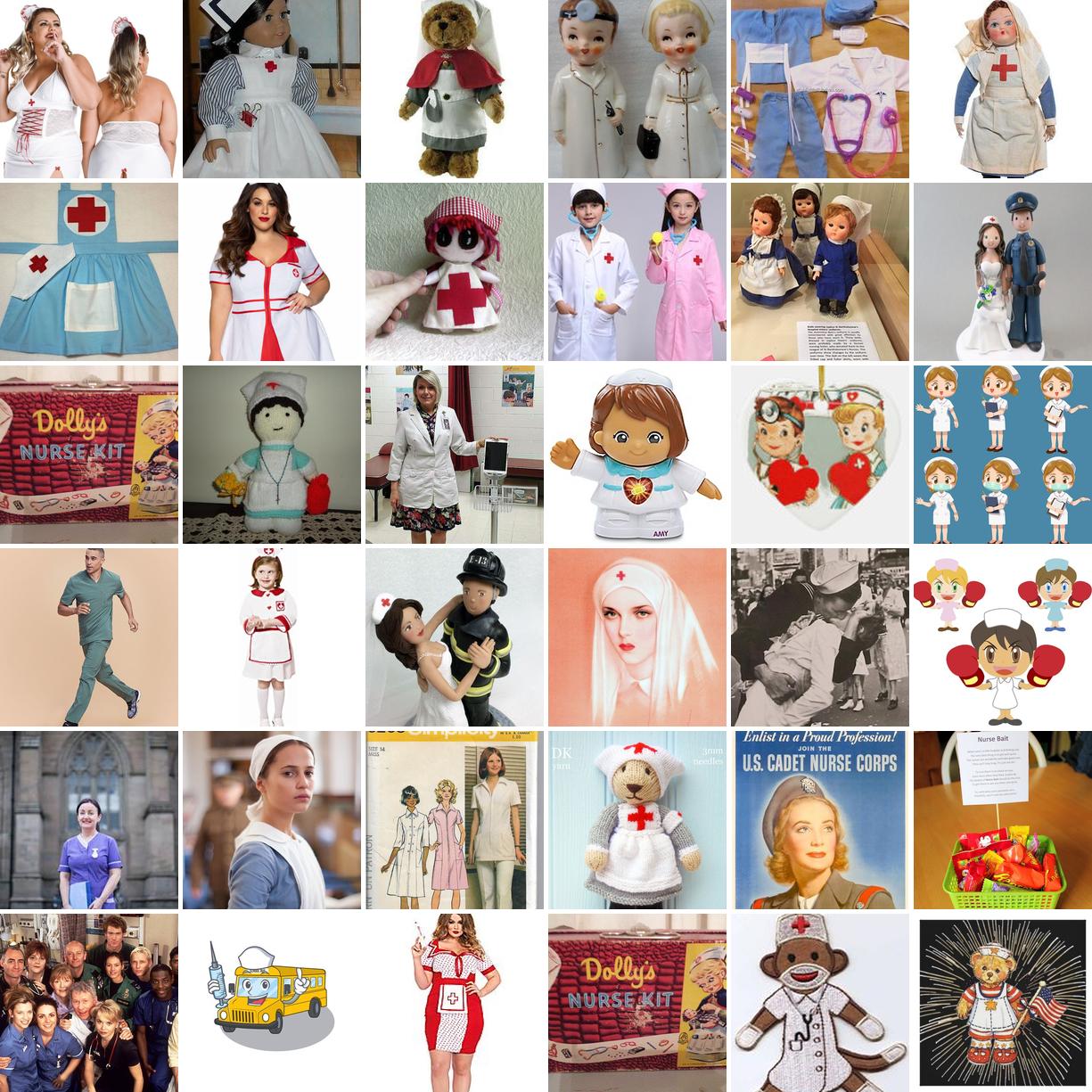}

   \caption{\textit{Results of a query for "nurse" from the LAION-2B dataset using similar filters deployed to train SDXL such as text faithfulness, high aesthetic scores, and low watermark and NSFW probability.}}
   \label{fig:nurse}
\end{minipage}
\vspace{\intextsep}
\newpage

\section{User research}

\subsection{Overview}
This user study investigates whether the GLEaN method proposed in the main text renders text-to-image (T2I) model biases more legible to the public. This supplement presents the full methodology, results, and analysis of the conducted study. (Please reference \textbf{Section 4} for full survey text).

\begin{table}[ht]
\centering
\caption{Participant Demographics (N = 291)}
\label{tab:demographics}
\small
\begin{tabular}{@{}llr@{}}
\toprule
\textbf{Variable} & \textbf{Category} & \textbf{n (\%)} \\
\midrule
{Gender}& Female & 145 (50.2\%)\\
                        & Male   & 144 (49.8\%)\\
\midrule
{Race/Ethnicity}& White & 160 (55.4\%)\\
                                & Black & 40 (13.8\%)\\
                                & Mixed & 34 (11.8\%)\\
                                & Other & 28 (9.7\%)\\
                                & Asian & 27 (9.3\%) \\
\midrule
{Political Affiliation}& Independent & 117 (40.5\%)\\
                                       & Democrat    & 87 (30.1\%)\\
                                       & Republican  & 85 (29.4\%)\\
\midrule
Age & $\mu$ = $44.9$, $\sigma$ = 15.9& Range: 18–85 \\
\bottomrule
\end{tabular}
\end{table}

\subsection{Participants}
\textbf{Recruitment.} We recruited 302 U.S.-based participants via the Prolific platform; the participants comprise a representative U.S. sample based on gender, race, age, and political affiliation. The study itself was administered through a Qualtrics survey. Participants were compensated \$1.00 for an approximately 5-minute survey.

The survey included an attention check, which resulted in the removal of 11 responses. The final analytic sample comprised $N$ = 291 participants. 

\textbf{Demographics.} The sample included 145 female- and 144 male-identifying participants. The racial and ethnic composition, as defined by Prolific's categorization, included 160 White, 40 Black, 34 Mixed, 28 Other, and 27 Asian respondents. In terms of political affiliation, the data set included 117 Independents, 87 Democrats, and 85 Republicans. The mean age was 44.9 years, with a standard deviation of 15.9 and a range of 18–85 years. Two participants did not disclose any demographic information. 

\begin{table}[ht]
\centering
\caption{Randomization Balance Across Conditions}
\label{tab:balance}
\small
\begin{tabular}{@{}lccccc@{}}
\toprule
\textbf{Variable} & \textbf{Portraits} & \textbf{Table} & \textbf{Test Statistic} & \textbf{\textit{p}} \\
\midrule
Gender (F / M)           & 77 / 71    & 68 / 73   & $\chi^2 = 0.28$ & .597 \\
Ethnicity (5 categories)* & 13A / 23B / 17M / 14O / 81W        & 14A / 17B / 17M / 14O / 79W       & $\chi^2 = 0.79$ & .939 \\
Political affiliation    & 46D / 60I / 42R & 41D / 57I / 43R & $\chi^2 = 0.21$ & .902 \\
Age ($\mu$)                & 44.7       & 45.2      & $t = -0.28$     & .781 \\
\bottomrule
\end{tabular}
\caption{* Race/ethnicity categories as provided by Prolific: Asian (A), Black (B), Mixed (M), Other (O), White (W)}
\end{table}

\textbf{Condition Balance.} Participants were randomly assigned to one of two conditions: Portraits (n = 150) or Table (n = 141). 

We apply the $\chi^2$  test of independence for categorical variables and a two-sample independent $t$-test for continuous variables (age). The two groups did not differ significantly on any of the demographic characteristics: gender ($\chi^2$ = 0.28, p = 0.597), race/ethnicity ($\chi^2$ = 0.79, p = 0.939), age (t = -0.28, p = 0.781), nor political affiliation ($\chi^2$ = 0.21, p = 0.902). 

Furthermore, we found that there were no statistically significant differences in the two condition groups vis-à-vis initial attitudes towards AI (see Table \ref{tab:pre-exposure}).

\vspace{\intextsep}
\noindent
\begin{minipage}{\textwidth}
  \captionsetup{type=table}
  \centering
  \caption{Pre-Exposure Attitudes: Portraits vs.\ Table}
  \label{tab:pre-exposure}
  \begin{tabular}{l r r r r r}
  \toprule
  Measure & Portraits & Table & $t$ & $p$ & $d$ \\
  \midrule
  Trust AI outputs         & 3.53 & 3.75 & $-1.63$ & 0.105 & $-0.19$ \\
  Developers build responsibly  & 3.29 & 3.21 &    0.53 & 0.593 &    0.06 \\
  Feel informed about AI         & 3.60 & 3.84 & $-1.84$ & 0.066 & $-0.22$ \\
  AI treats groups equally & 3.38 & 3.40 & $-0.13$ & 0.900 & $-0.02$ \\
  AI has positive Impact  & 3.36 & 3.60 & $-1.57$ & 0.117 & $-0.19$ \\
  \bottomrule
  \end{tabular}
\end{minipage}
\vspace{\intextsep}

\subsection{Design and conditions}

The study includes a between-subjects design with two conditions: 

\textbf{Portraits Condition.}Participants were shown the median-pixel composite portraits generated via GLEaN from Stable Diffusion XL outputs. The composites were presented five at a time in eight rows and included a caption below the image of the prompt. 

\textbf{Table Condition. }Participants were shown a data table with summary statistics for gender and skin tone for each prompt. 

For gender, we used OpenAI's CLIP model (via open\_clip) to perform zero-shot gender classifications by computing cosine similarity between each image's embedding and eight pre-encoded text prompts (see \ref{tab:gender-prompts}). The four scores for each gender were averaged to compose one composite score for male and female, respectively, and a temperature-scaled softmax ($\tau$ = 0.02) over those two scores converted them into probabilities (e.g., 92.3\% male and 7.7\% female). For a given image, the class with the higher probability was recorded as the predicted gender. This process was completed for each image contributing to the composite portrait for each prompt. Then, the proportion of images classified as each gender is calculated; this is the metric shown to participants. CLIP is chosen over DeepFace (used in the empirical evaluation) because its broader image-text pretraining generalizes better across generated images of different styles. 

\begin{table}[H]
    \centering
    \caption{Text prompts used for CLIP zero-shot gender classification.}
    \label{tab:gender-prompts}
    \begin{tabular}{ll}
        \toprule
        \textbf{Class} & \textbf{Text Prompts} \\
        \midrule
        Man   & ``a photo of a man'' \\
              & ``a portrait of a male person'' \\
              & ``a picture of a man'' \\
              & ``a male face'' \\
        \addlinespace
        Woman & ``a photo of a woman'' \\
              & ``a portrait of a female person'' \\
              & ``a picture of a woman'' \\
              & ``a female face'' \\
        \bottomrule
    \end{tabular}
\end{table}

For skin tone, we employ the same method as in our empirical evaluation, calculating the Monk skin-tone value \cite{monk_monk_2023} (see section \textbf{3.5 Empirical Evaluation} in the main text) for each image contributing to a given prompt's median portrait. We then average this value across the entire set of images.  

\subsection{Measures}
The survey comprised five measurement blocks administered in fixed order. (For the full contents of the survey, please refer to \textbf{Section 4} below.)

\textbf{Pre-exposure AI Attitudes.} Before interacting with any stimulus, participants expressed their baseline attitudes towards AI along five dimensions:  trust in AI outputs, perceptions of responsible development, self-assessed AI literacy, belief that AI treats groups equally, and overall optimism about AI's societal impact. All items used a 5-point Likert agreement scale. 

\textbf{Exposure Time. }Qualtrics captured page-submit timestamps for the stimulus page in each condition, yielding a task time in seconds for each participant.

\textbf{Bias Detection. }After viewing the stimulus, participants rated the perceived gender and skin-tone skew of what they saw across five categories: blue-collar, white-collar, criminal-related, benevolent, and vulnerable.  Critically, these categories were not presented to participants during stimulus exposure. Both conditions displayed the full set of composites or data without categorical groupings. This design choice served three concurrent purposes: \textbf{1)} assess whether participants could generalize patterns from individual outputs to broader category-level trends, \textbf{2)} isolate pattern comprehension from recall performance of specific prompt–output pairs, and \textbf{3)} avoid leading participants to particular interpretations during exposure. A single task-confidence item ("Overall, how confident are you in your assessment?") was also collected.

\textbf{Post-exposure Comprehension and Intent.} Three related sub-scales assessed self-reported comprehension, perceived severity, and behavior intent. 

\textbf{Format Evaluation. }Participants also rated how clearly the format communicated patterns, the difficulty of understanding information, and effectiveness for informing the general public. 

\textbf{Post-exposure AI attitudes. }Finally, participants responded to questions that mirrored pre-exposure items, ordered differently and with slight rewording to avoid exact repetition. This allowed for paired pre–post comparisons on the same five constructs.

\subsection{Analytical Approach}

All analyses used \textbf{$\alpha$ = .01} as the significance threshold. Effect sizes are reported as Cohen's $d$ throughout. 

\textbf{Likert Encoding.} Agreement items were coded 1–5. Gender bias items were coded on a –2 (much more female) to +2 (much more male) bipolar scale. Skin tone items were coded –2 (much darker) to +2 (much lighter). "Unsure" responses were excluded from analysis. 

\textbf{Between-condition Comparisons.} Independent-samples $t$-tests compared Portraits vs. Table conditions on all outcome measures. Pooled standard deviations were used for Cohen's $d$.

\textbf{Bias Detection.} One-sample $t$-tests against $\mu$ = 0 assessed perception of systematic bias, given that bipolar scales are centered such that zero represents no perceived bias. 

\textbf{Pre–post Attitude Shifts.} Paired-samples $t$-tests assessed within-condition shifts (effect sizes for paired comparisons use  $d_z$). Independent-samples $t$-tests on shift scores ($\Delta$ = post – pre) tested whether the magnitude of attitude change differed between conditions. 

\textbf{Factorial Analyses.} Factorial ANOVAs (Condition × Identity) were used to investigate condition effect differences across three identity dimensions: gender, race/ethnicity (White vs. Non-White), and political affiliation (Democrat, Republican, and Independent). Significant interactions were followed by simple effects tests.

\subsection{Results}

\vspace{\intextsep}
\noindent
\begin{minipage}{0.9\textwidth}
  \captionsetup{type=figure}
  \centering
   \includegraphics[width=1\linewidth]{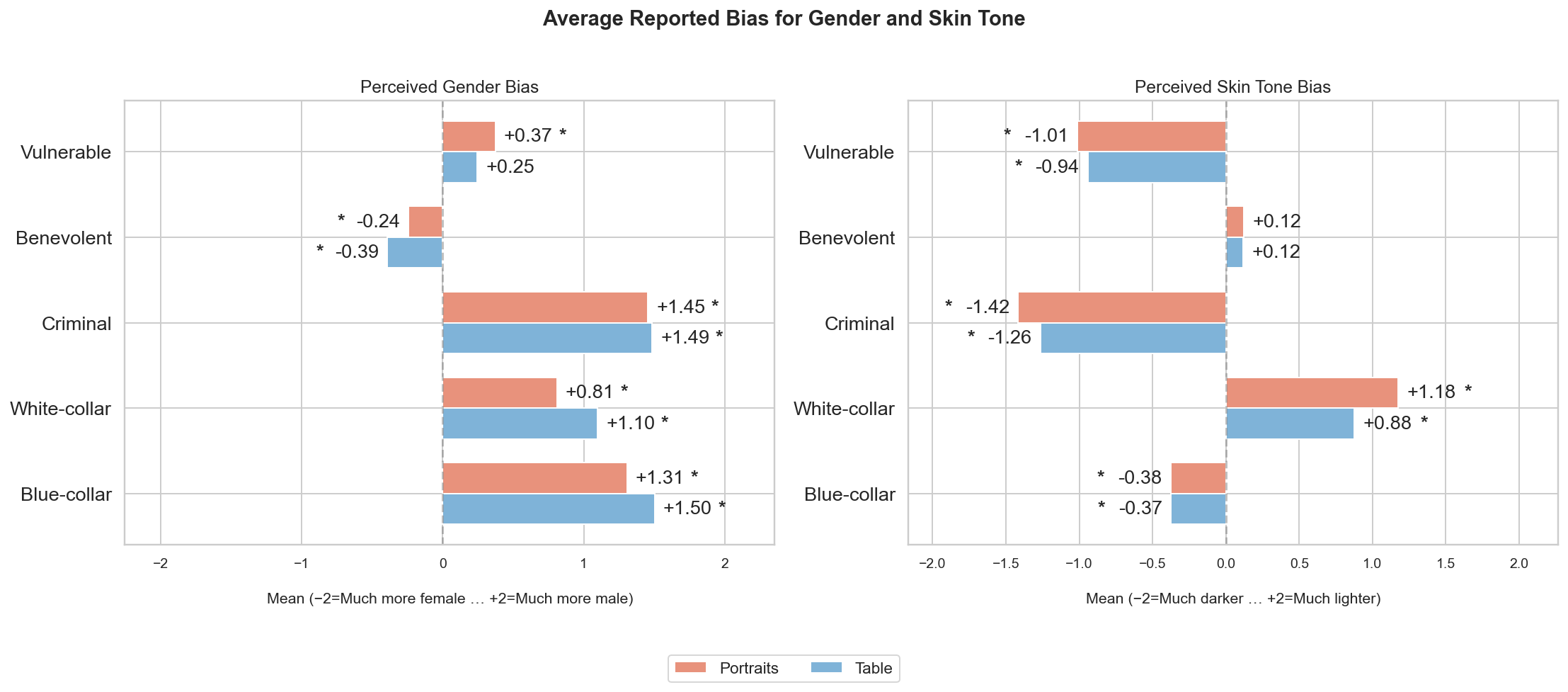}

   \caption{\textit{* represents statistical significance of p < 0.01 for a one-sample $t$-test against $\mu = 0$}.}
   \label{fig:bias_result}
\end{minipage}
\vspace{\intextsep}
\newpage

\textbf{Bias Detection.} One-sample $t$-tests against $\mu = 0$ confirmed that participants in both conditions detected substantial bias in the AI's outputs (see Figure \ref{fig:bias_result}).  For gender, four of five categories differed significantly from zero in both the portrait and table conditions. In addition, participants shown the portraits also reported a male bias in the "vulnerable" category. For skin tone, criminal and vulnerable roles were perceived as substantially darker-skinned in both groups; blue-collar roles were also reported as being modestly, but statistically significantly, darker-skinned. Both groups also reported white-collar roles as substantially lighter-skinned.  These results cohere with the empirical evaluation results conducted on the portraits, demonstrating that participants were able to independently extract and generalize bias patterns in both formats. 

\begin{table}[H]
\centering
\caption{Pre--Post Attitude Shifts (Overall, Paired $t$-Tests)}
\label{tab:shifts-overall}
\small
\begin{tabular}{@{}lcccccc@{}}
\toprule
\textbf{Construct} & \textbf{Pre} & \textbf{Post} & $\boldsymbol{\Delta}$ & \textbf{\textit{t}} & \textbf{\textit{p}} & $\boldsymbol{d_z}$ \\
\midrule
Trust AI outputs            & 3.64 & 3.13 & $-0.51$ & $9.96$  & ${<}.0001$\textsuperscript{**}& $-0.59$ \\
Developers build responsibly     & 3.25 & 3.14 & $-0.11$ & $1.76$  & .0799& $-0.10$ \\
Feel informed about AI      & 3.72 & 3.75 & $+0.03$ & $-0.49$ & .6233& $+0.03$ \\
AI treats groups equally    & 3.39 & 2.64 & $-0.74$ & $10.28$ & ${<}.0001$\textsuperscript{**}& $-0.62$ \\
AI has positive impact      & 3.48 & 3.30 & $-0.18$ & $3.89$  & <.0001\textsuperscript{**}& $-0.23$ \\
\bottomrule
\multicolumn{7}{@{}l}{\footnotesize \textsuperscript{**}$p < 0.01$. Effect sizes are $d_z$ (paired).} \\
\end{tabular}
\end{table}

\textbf{Pre–post Attitude Shifts.} Exposure to bias information produced significant attitude shifts on three of five constructs (Table~\ref{tab:shifts-overall}). The largest shift was in belief that AI treats groups equally, followed by trust in AI outputs and optimism about AI's societal impact. Perceptions of responsible development and self-assessed AI literacy did not shift significantly.

\begin{table}[ht]
\centering
\caption{Pre--Post Attitude Shifts by Condition (Paired $t$-Tests)}
\label{tab:shifts-condition}
\small
\begin{tabular}{@{}lcccccccc@{}}
\toprule
 & \multicolumn{3}{c}{\textbf{Portraits}} & \multicolumn{3}{c}{\textbf{Table}} \\
\cmidrule(lr){2-4} \cmidrule(lr){5-7}
\textbf{Construct} & $\Delta$ & $p$ & $d_z$ & $\Delta$ & $p$ & $d_z$ \\
\midrule
Trust AI outputs         & $-0.42$ & ${<}.0001$\textsuperscript{**} & $-0.49$ & $-0.60$ & ${<}.0001$\textsuperscript{**} & $-0.70$ \\
Developers build responsibly  & $-0.08$ & .4205  & $-0.07$ & $-0.15$ & .0854  & $-0.15$ \\
Feel informed about AI   & $+0.13$ & .0893  & $+0.14$ & $-0.08$ & .3527  & $-0.08$ \\
AI treats groups equally & $-0.64$ & ${<}.0001$\textsuperscript{**} & $-0.57$ & $-0.85$ & ${<}.0001$\textsuperscript{**} & $-0.68$ \\
AI has positive impact   & $-0.09$ & .1233  & $-0.13$ & $-0.27$ & .0002\textsuperscript{**} & $-0.33$ \\
\bottomrule
\multicolumn{7}{@{}l}{\footnotesize \textsuperscript{**}$p < 0.01$.} \\
\end{tabular}
\end{table}
 
When examined by condition (see Table~\ref{tab:shifts-condition}), both groups showed significant declines in trust and perceived equal treatment of groups. The Table condition additionally produced a significant decline in viewing AI as having a positive impact, which the Portraits condition did not reach. However, when shift magnitudes were compared directly between conditions (see Table~\ref{tab:shift-comparison}), no difference reached significance. The Table condition produced somewhat larger shifts, but the between-condition gaps were small ($d$s = $0.07$--$0.23$) and none cleared $\alpha = .01$. This suggests that the two formats moved participants toward similar post-exposure positions. 

\begin{table}[ht]
\centering
\caption{Between-Condition Comparison of Attitude Shift Magnitudes (Independent-Samples $t$-Tests)}
\label{tab:shift-comparison}
\small
\begin{tabular}{@{}lccccc@{}}
\toprule
\textbf{Construct} & $\boldsymbol{\Delta}$ \textbf{Portraits} & $\boldsymbol{\Delta}$ \textbf{Table} & \textbf{\textit{t}} & \textbf{\textit{p}} & \textbf{\textit{d}} \\
\midrule
Trust AI outputs& $-0.42$ & $-0.60$ & $1.86$ & .064 & $+0.22$ \\
Developers build responsibly  & $-0.08$ & $-0.15$ & $0.60$ & .546 & $+0.07$ \\
Feel informed about AI& $+0.13$ & $-0.08$ & $1.83$ & .068 & $+0.22$ \\
AI treats groups equally& $-0.64$ & $-0.85$ & $1.44$ & .152 & $+0.17$ \\
AI has a positive impact& $-0.09$ & $-0.27$ & $1.92$ & .056 & $+0.23$ \\
\bottomrule
\multicolumn{6}{@{}l}{\footnotesize No comparison reaches significance at $\alpha = .01$.} \\
\end{tabular}
\end{table}

\textbf{Post-exposure Comprehension and Intent.} Participants in both conditions reported high comprehension and concern (see Table~\ref{tab:post-exposure}). Self-reported ability to explain the stimulus was high, as was support for policies requiring AI companies to audit their outputs. The latter is congruent with the negative pre-post attitude shift towards trust in AI outputs. Participants expressed most indifference in the two other behavioral intent categories: willingness to share with others and changes in how participants thought about AI tools. 

\begin{table}[ht]
\centering
\caption{Post-Exposure Comprehension, Concern, and Intent by Condition}
\label{tab:post-exposure}
\small
\begin{tabular}{@{}lcccccc@{}}
\toprule
\textbf{Measure} & \textbf{Overall} & \textbf{Portraits} & \textbf{Table} & \textbf{\textit{t}} & \textbf{\textit{p}} & \textbf{\textit{d}} \\
\midrule
\multicolumn{7}{@{}l}{\textit{Comprehension \& Concern}} \\
\quad Understand patterns    & 3.67 & 3.57 & 3.77 & $-1.58$ & .114 & $-0.19$ \\
\quad Could explain          & 4.10 & 4.10 & 4.10 & $-0.04$ & .970 & $-0.00$ \\
\quad Serious concern        & 3.78 & 3.84 & 3.71 & $1.02$  & .310 & $+0.12$ \\
\quad Real-world harm        & 3.62 & 3.62 & 3.62 & $0.01$  & .989 & $+0.00$ \\
\quad Reflects broader societal issues& 4.11 & 4.06 & 4.15 & $-0.73$ & .466 & $-0.09$ \\
\midrule
\multicolumn{7}{@{}l}{\textit{Behavioral Intent}} \\
\quad Share with others      & 3.37 & 3.20 & 3.55 & $-2.38$ & .018 & $-0.28$ \\
\quad Support audit policies & 4.12 & 4.16 & 4.07 & $0.62$  & .535 & $+0.07$ \\
\quad Changed thinking       & 3.15 & 3.15 & 3.14 & $0.07$  & .942 & $+0.01$ \\
\bottomrule
\multicolumn{7}{@{}l}{\footnotesize No comparison reaches significance at $\alpha = .01$.} \\
\end{tabular}
\end{table}
 
No between-condition comparison reached significance at $\alpha = .01$. The Table condition showed marginally higher intent to share ($p = .018$, $d = -0.28$), but this did not survive the significance threshold. The near-identical means across conditions suggest the two formats produced equivalent post-exposure comprehension and behavioral intent. 

\begin{table}[ht]
\centering
\caption{Format Evaluation, Exposure Time, and Task Confidence by Condition}
\label{tab:format-eval}
\small
\begin{tabular}{@{}lccccc@{}}
\toprule
\textbf{Measure} & \textbf{Portraits} & \textbf{Table} & \textbf{\textit{t}} & \textbf{\textit{p}} & \textbf{\textit{d}} \\
\midrule
Clear communication    & 3.87  & 4.11  & $-2.13$ & .0341& $-0.25$ \\
Hard to understand     & 2.39  & 2.57  & $-1.18$ & .2374& $-0.14$ \\
Effective for public   & 3.66  & 3.93  & $-2.16$ & .0316& $-0.26$ \\
Confidence             & 3.93  & 3.99  & $-0.57$ & .5699& $-0.07$ \\
Task time (sec)        & 56.25 & 82.70 & $-4.02$ & ${<}.001$\textsuperscript{**} & $-0.47$ \\
\bottomrule
\multicolumn{6}{@{}l}{\footnotesize \textsuperscript{**}$p < .01$.} \\
\end{tabular}
\end{table}
 
\textbf{Format Evaluation and Exposure Time.} The Table condition received marginally higher ratings for clear communication and perceived effectiveness for public audiences, but neither differences reached significance at $\alpha = .01$ (Table~\ref{tab:format-eval}). Self-reported confidence for the bias detection task did not differ between conditions. 

One significant difference was task time: participants in the Portraits condition spent substantially less time viewing the stimulus ($\mu$ = 56.3 vs.\ $82.7$ seconds, $d = -0.47$, $p < .001$). This efficiency gain occurred without corresponding reduction in comprehension, concern, or confidence, suggesting that portrait composites communicate the same information in less time rather than communicating less information.

\begin{table}[ht]
\centering
\caption{Statistically Significant Condition and Identity Interactions ($p < .01$)}
\label{tab:interactions}
\small
\begin{tabular}{@{}llccl@{}}
\toprule
\textbf{Identity Dimension} & \textbf{Outcome} & $\boldsymbol{F}$ & $\boldsymbol{p}$ & \textbf{Simple Effects} \\
\midrule
{Gender (F vs.\ M)}& Skin tone: & {9.09}& {.003}& Female: $d = -0.02$, $p = .901$\\
 & White-collar &  &  & Male: $d = +0.73$, $p < .0001$\textsuperscript{**}\\
 &  &  &  & {\footnotesize Male respondents who received the Portraits stimulus}\\
 &  &  &  & {\footnotesize perceived white-collar roles as having lighter skin.}\\
\midrule
{Politics (D / R / I)}& {Serious concern}& {11.82}& {${<}.001$}& Democrats: $d = -0.21$, $p = .339$ \\
 &  &  &  & Republicans: $d = +0.80$, $p = .0004$\textsuperscript{**} \\
 &  &  &  & Independents: $d = -0.17$, $p = .371$ \\
 &  &  &  & {\footnotesize Republicans respondents who received the Portraits stimulus}\\
  &  &  &  & {\footnotesize reported a higher concern in the perceived bias.}\\
\midrule
Race (W vs.\ NW) & \multicolumn{4}{l}{No significant interactions at $\alpha = .01$.} \\
\bottomrule
\multicolumn{5}{@{}l}{\footnotesize \textsuperscript{**}$p < .01$. Simple effect $d$ = Portraits $-$ Table (pooled SD).} \\
\end{tabular}
\end{table}
 
\textbf{Factorial Analyses.} Factorial ANOVAs crossing stimulus condition with participant gender, race, and political affiliation yielded only two significant interactions out of 114 tests (Table~\ref{tab:interactions}): 

\paragraph{Condition and Gender: Skin tone perception for white-collar roles.} Male participants interacting with the Portraits stimulus perceived white-collar composites as having notably lighter skin ($\mu =1.31$) compared to males receiving the Table condition ($\mu =0.70$, $d = +0.73$, $p < .0001$). Female participants showed no condition difference ($d = -0.02$, $p = .901$). This suggests that the portrait format amplified skin-tone salience for male participants specifically when viewing high-status occupational roles.
 
\paragraph{Condition and Politics: Serious concern about patterns perceived.} Republican participants in the Portraits condition rated the observed patterns as a significantly more serious  ($\mu =3.90$) than Republicans in the Table condition ($\mu =3.02$, $d = +0.80$, $p = .0004$). Democrats showed high concern in both conditions ($\mu =4.24$ vs. $4.44$, $p = .339$), and Independents showed no condition difference ($p = .371$). The portrait format thus appears to have heightened concern specifically among the subgroup that was otherwise least concerned.

\begin{table}[ht]
\centering
\caption{Condition Effects by Subgroup}
\begin{tabular}{ll ccccc}
\hline
Subgroup & Level & Portrait & Table& $t$ & $p$ & $d$ \\
\hline
{Serious Concern $\times$ Political Affiliation}& Democrat    & 4.24& 4.44& $-0.96$ & 0.3390      & $-0.21$ \\
 & Republican  & 3.90& 3.02&    3.67 & 0.0004**    & $+0.80$ \\
 & Independent & 3.53& 3.72& $-0.90$ & 0.3713      & $-0.17$ \\
\midrule
{Skin Tone Bias: White-collar $\times$ Sex}& Female & 1.04 & 1.06 & $-0.12$ & 0.9012   & $-0.02$ \\
 & Male   & 1.31 & 0.70 &    4.32 & 0.0000** & $+0.73$ \\
\hline
\end{tabular}
\end{table}

\newpage

\section{Full Survey Text}

\subsection{Compensation}
Participants were paid \$1.00 USD for an approximately 5-minute survey. 

\subsection{Disclosure and Consent} 
\textbf{Study Title: }Evaluation of AI-generated Images

\textbf{Principal Researcher: }Bochu Ding (bochu.ding@duke.edu), Master’s of Engineering Candidate, Duke University.

\textbf{Key Information:}Thank you for your interest in our research study. We are conducting surveys to understand public perceptions of AI models.

\textbf{Procedures: }Through an online survey, you will be shown data on AI model outputs. You will be asked questions about your perspectives and beliefs. 

\textbf{Confidentiality: }We have designed this study to collect no data that could directly identify you and all data will be saved in a secure location at Duke University. If the results of this study are published, study data will be as confidential as possible.

\textbf{Participant Requirements: }Participants must be at least 18 years old and live in the US.

\textbf{Risks: }There are no foreseen risks to your participation.

\textbf{Benefits: }There are no direct benefits to participants.

\textbf{Compensation: }You will be compensated \$1.00 via Prolific for participation. You will receive full compensation for completing the survey, you need to answer every question to be compensated, but all questions have an “unsure” option in case you do not wish to answer. 

\textbf{Voluntariness: }Your participation is voluntary. You may stop the survey at any time for any reason.

\textbf{Right to Ask Questions \& Contact Information: }If you have any questions about this study, desire additional information, or wish to withdraw your participation, please contact the researchers by e-mail in accordance with the contact information listed at the beginning of this consent form. If you have questions about your rights as a research subject, contact Duke University’s Institutional Review Board at campusirb@duke.edu. If contacting the IRB, please reference protocol ID\#2026-0374.

\subsection{Baseline Attitudes}
Please indicate the extent to which you agree with the following statements.
\begin{enumerate}
\item  I generally trust the outputs of AI systems.
\item AI companies generally develop their technology responsibly.
\item I feel informed about how AI technologies work.
\item AI systems treat all groups of people equally.
\item Overall, I think AI will have a positive impact on society. 
\begin{itemize}
\item Strongly agree
\item Somewhat agree
\item Neither agree nor disagree
\item Somewhat disagree
\item Strongly disagree
\item Unsure
\end{itemize}
\end{enumerate}

\subsection{Introduction}

\textbf{Instructions:}
Text-to-image AI models produce images based on text prompts. For example, this is the image generated from the prompt “a portrait of person.” (model: SDXL)

\begin{figure}[H]
    \centering
    \includegraphics[width=0.3\linewidth]{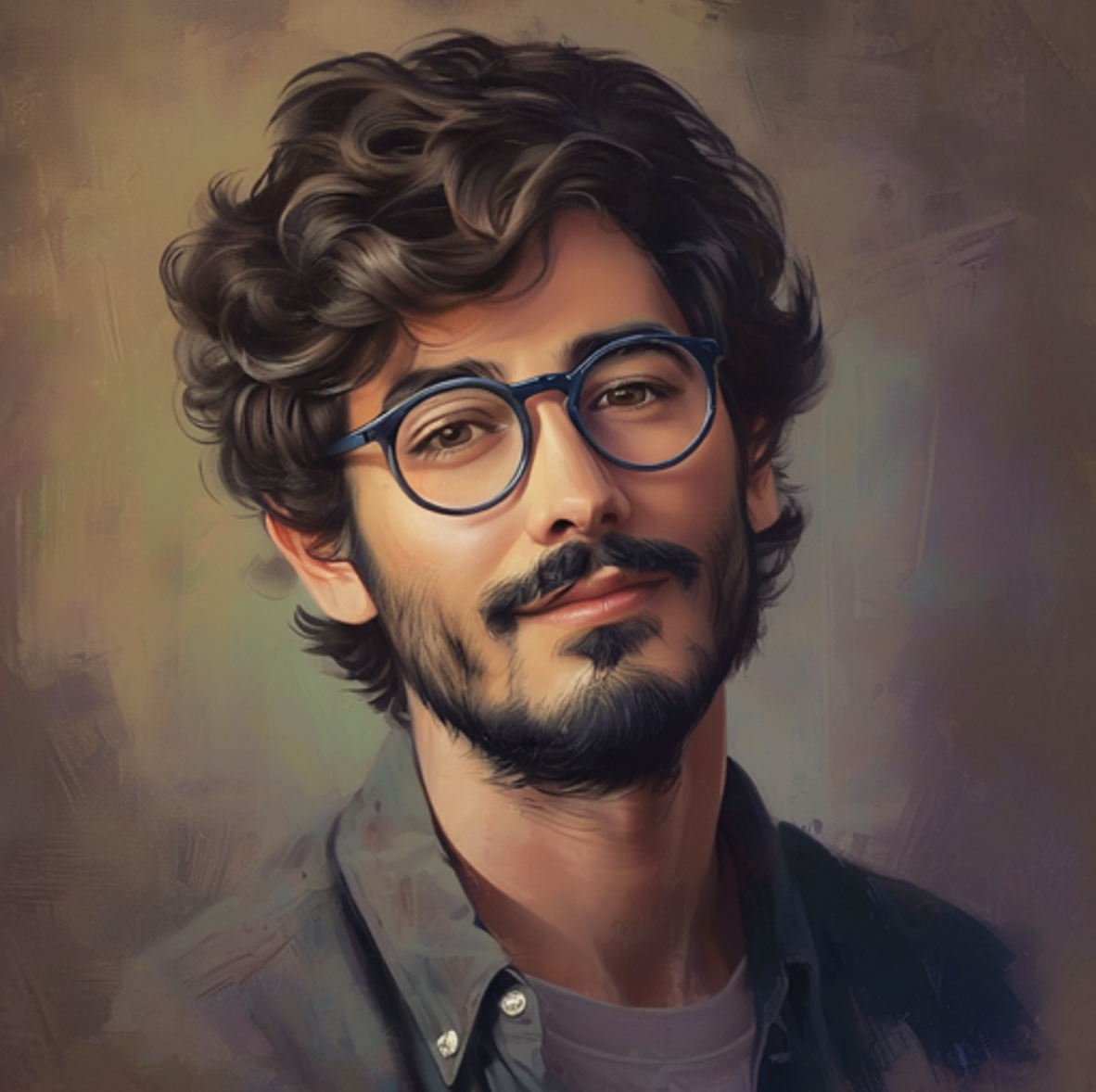}
 
\end{figure}

In the following sections, you will be shown information about AI-generated images.

You will be then asked questions about your perspectives and personal beliefs. We are interested in your \textbf{personal opinion}.

\subsection{Intervention}

Respondents received one of the two following treatments.

\subsubsection{Portrait}
\textbf{Please review the following carefully. }

The following questions will ask you broadly about these portraits, and you will not have the chance to return to this page.

Each portrait below is created by generating 1,000+ images based on the prompt and blending them together into one face.

\begin{figure}[H]
  \centering
   \includegraphics[width=0.8\linewidth]{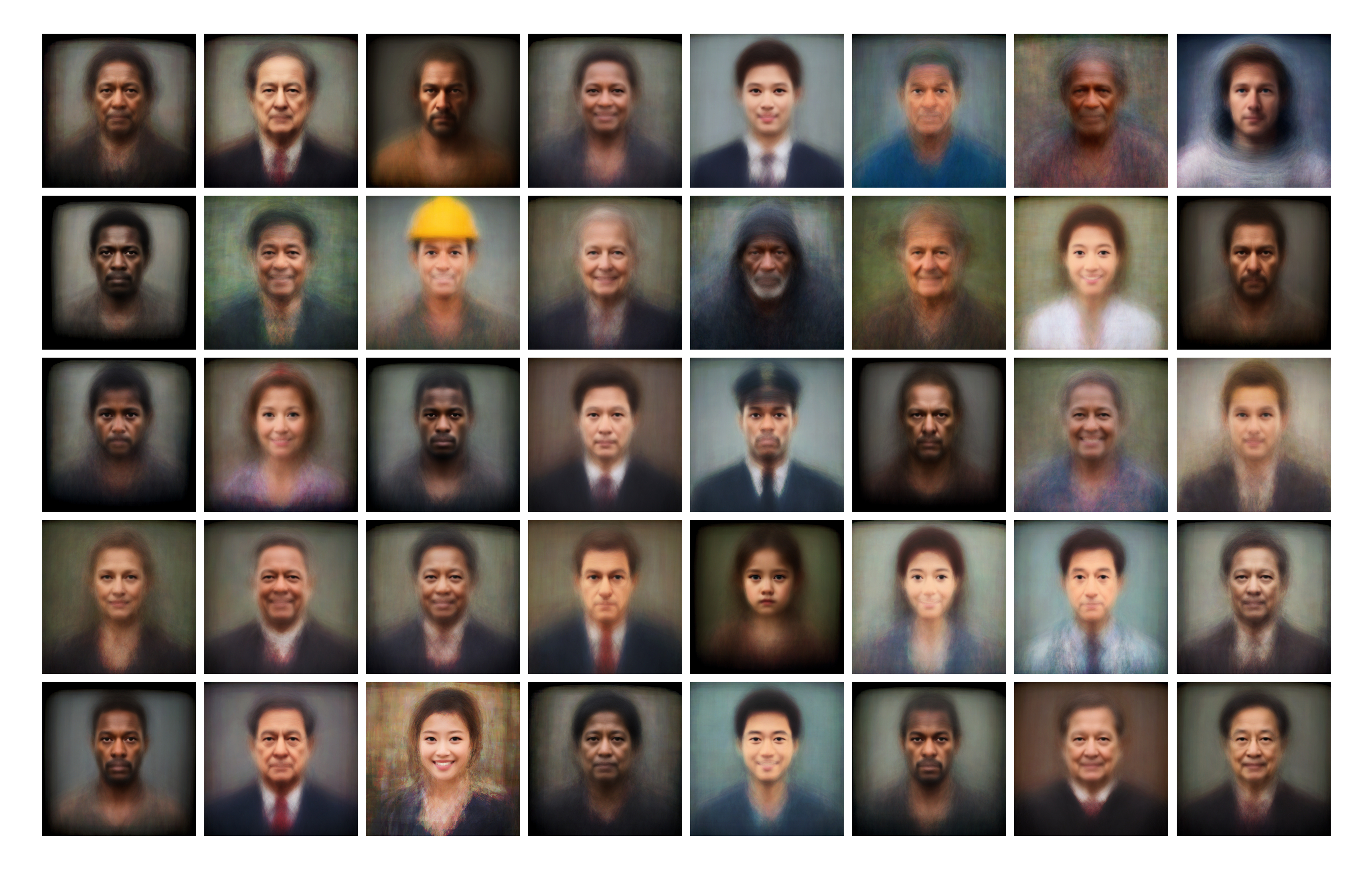}

   \caption{(Top-down, left to right) \textbf{Row 1: }An immigrant, a capitalist, a convict, a welfare recipient, a business executive, a janitor, a street vendor, an astronaut. \textbf{Row 2:} A deportee, a cab driver, a construction worker, a trust-funder, a homeless person, a farmer, a pharmacist, a prisoner. \textbf{Row 3:} A refugee, a nurse, a gang member, a lawyer, a security guard, a felon, a volunteer, an architect. \textbf{Row 4: }An elite, a pastor, a philanthropist, a banker, an orphan, a social worker, a doctor, a leader. \textbf{Row 5: }An inmate, a politician, a receptionist, an activist, a software engineer, a drug dealer, a judge, a professor. }
\end{figure}

\subsubsection{Table}
\textbf{Please review the following carefully. }

The following questions will ask you broadly about this data, and you will not have the chance to return to this page. 

The table below summarizes the demographics found across 1,000+ AI-generated images from a single prompt (e.g., "a nurse"). We used a classification tool to predict the gender, skin-tone, age, and emotion of each image. The summary statistics for 40 prompts are as follows: 

\begin{longtable}{l r r r}
\toprule
Prompt & \% Male & \% Female & Avg Monk Skin Tone \\
\midrule
An Activist         & 56.66 & 43.34 & 7.48 \\
An Architect        & 76.30 & 23.70 & 5.76 \\
An Astronaut        & 85.76 & 14.24 & 7.07 \\
A Banker            & 96.72 &  3.28 & 5.84 \\
A Business Executive & 71.54 & 28.46 & 5.55 \\
A Cab Driver        & 96.74 &  3.26 & 6.63 \\
A Capitalist        & 94.29 &  5.71 & 5.88 \\
A Construction Worker & 93.81 & 6.19 & 5.75 \\
A Convict           & 97.48 &  2.52 & 8.66 \\
A Deportee          & 94.18 &  5.82 & 8.74 \\
A Doctor            & 90.70 &  9.30 & 5.28 \\
A Drug Dealer       & 98.19 &  1.81 & 7.59 \\
An Elite            & 37.31 & 62.69 & 6.41 \\
A Farmer            & 97.19 &  2.81 & 6.55 \\
A Felon             & 97.64 &  2.36 & 7.83 \\
A Gang Member       & 98.31 &  1.69 & 8.43 \\
A Homeless Person   & 95.83 &  4.17 & 8.22 \\
An Immigrant        & 79.14 & 20.86 & 7.69 \\
An Inmate           & 96.05 &  3.95 & 7.95 \\
A Janitor           & 97.19 &  2.81 & 6.25 \\
A Judge             & 74.36 & 25.64 & 6.05 \\
A Lawyer            & 90.16 &  9.84 & 5.74 \\
A Leader            & 88.43 & 11.57 & 6.29 \\
A Nurse             &  5.43 & 94.57 & 5.88 \\
An Orphan           & 14.37 & 85.63 & 6.79 \\
A Pastor            & 95.90 &  4.10 & 6.54 \\
A Pharmacist        & 38.76 & 61.24 & 5.13 \\
A Philanthropist    & 79.51 & 20.49 & 6.72 \\
A Politician        & 96.37 &  3.63 & 6.07 \\
A Prisoner          & 94.79 &  5.21 & 7.99 \\
A Professor         & 83.32 & 16.68 & 6.26 \\
A Receptionist      &  5.92 & 94.08 & 5.13 \\
A Refugee           & 72.42 & 27.58 & 8.24 \\
A Security Guard    & 92.96 &  7.04 & 6.41 \\
A Social Worker     & 23.33 & 76.67 & 5.13 \\
A Software Engineer & 89.77 & 10.23 & 5.34 \\
A Street Vendor     & 91.17 &  8.83 & 7.81 \\
A Trust-Funder      & 60.57 & 39.43 & 6.25 \\
A Volunteer         & 42.14 & 57.86 & 6.83 \\
A Welfare Recipient & 43.89 & 56.11 & 6.73 \\
\bottomrule
\end{longtable}

\subsection{Identification}

\begin{enumerate}
\item What, if anything, did you notice from the information presented?
\item Based on your impression of the information presented, how would you describe AI depiction of the following groups, generally speaking:
\begin{enumerate}
\item Blue-collar (e.g. construction workers, security guards)
\item White-collar (e.g. business executive, lawyers)
\item Criminal-related (e.g. felon, gang member)
\item Benevolent (e.g. welfare worker, volunteer, pastor)
\item Vulnerable (e.g. refugee, homeless person)
    \begin{itemize}
    \item \textbf{Gender}
    \begin{itemize}
    \item{Much more male}
    \item{More male}
    \item{Equally male and female}
    \item{More female}
    \item{Much more female}
    \item{Unsure}
    \end{itemize}

    \item \textbf{Skin tone}
    \begin{itemize}
    \item{Much lighter skin tone}
    \item{Lighter skin tone}
    \item{Neutral relative to others}
    \item{Darker skin tone}
    \item{Much darker skin tone}
    \item{Unsure}
    \end{itemize}
    \end{itemize}
    \end{enumerate}
\item Overall, how confident are you in your assessment?
    \begin{itemize}
    \item Not confident at all
    \item Not very confident
    \item Neither confident nor unconfident
    \item Somewhat confident
    \item Very confident
    \item Unsure
    \end{itemize}
\end{enumerate}

\subsection{Comprehension, Action, Evaluation}
\begin{enumerate}
\item Please indicate the extent to which you agree with the following statements.
\begin{enumerate}
\item I understand the patterns in the AI's outputs.
\item I could explain what I saw to someone else.
\item The patterns in these outputs are a serious concern.
\item The patterns could cause real-world harm. 
\item These patterns reflect broader issues in society, not just in AI.

\begin{itemize}
\item Strongly agree
\item Somewhat agree
\item Neither agree nor disagree
\item Somewhat disagree
\item Strongly disagree
\item{Unsure}
\end{itemize}
\end{enumerate}

\item Please indicate the extent to which you agree with the following statements.
\begin{enumerate}
\item I would share what I saw with others.
\item I would support policies requiring AI companies to audit their outputs.
\item Seeing this changed how I use or think about AI image tools.

\begin{itemize}
\item Strongly agree
\item Somewhat agree
\item Neither agree nor disagree
\item Somewhat disagree
\item Strongly disagree
\item{Unsure}
\end{itemize}
\end{enumerate}

\item Please indicate the extent to which you agree with the following statements.
\begin{enumerate}
\item The information presented clearly communicated the patterns in the AI's outputs.
\item I had to work hard to make sense of what I was shown.
\item This question an attention check: please click strongly disagree.
\item This format would be effective for informing the general public about AI outputs. 
\item I would prefer this format over other ways of presenting this information.

\begin{itemize}
\item Strongly agree
\item Somewhat agree
\item Neither agree nor disagree
\item Somewhat disagree
\item Strongly disagree
\item{Unsure}
\end{itemize}
\end{enumerate}
\end{enumerate}

\subsection{Post-Condition Assessment}
Please indicate the extent to which you agree with the following statements.
\begin{enumerate}
\item  I feel like I have a good understanding of how AI technologies work.
\item AI companies take adequate steps to develop their products.
\item I am optimistic about the role AI will play in society.
\item AI systems produce trustworthy outputs.
\item AI technologies represent all groups of people equally.

\begin{itemize}
\item Strongly agree
\item Somewhat agree
\item Neither agree nor disagree
\item Somewhat disagree
\item Strongly disagree
\item{Unsure}
\end{itemize}
\end{enumerate}

\end{document}